\documentclass[a4paper,fleqn]{cas-dc}

\usepackage{subcaption}
\usepackage[acronym]{glossaries} 
\usepackage[numbers]{natbib}
\usepackage{lipsum}

\newacronym{ml}{ML}{}
\newacronym{fl}{FL}{Federated Learning}
\newacronym{iot}{IoT}{Internet of Things}
\newacronym{smc}{SMC}{Secure Multiparty Computation}
\newacronym{eeg}{EEG}{Electroencephalogram}

\def\tsc#1{\csdef{#1}{\textsc{\lowercase{#1}}\xspace}}
\tsc{WGM}
\tsc{QE}
\tsc{EP}
\tsc{PMS}
\tsc{BEC}
\tsc{DE}

\begin{document}
\let\WriteBookmarks\relax
\def\floatpagepagefraction{1}
\def\textpagefraction{.001}

\shorttitle{Privacy-Preserving Edge Federated Learning}

\shortauthors{Amin Aminifar et~al.}

\title [mode = title]{Privacy-Preserving Edge Federated Learning for Intelligent Mobile-Health Systems}                      
\tnotemark[1]

\tnotetext[1]{This research has been partially supported by the Wallenberg AI, Autonomous Systems and Software Program (WASP), Swedish Research Council (VR), ELLIIT Strategic Research Environment, Swedish Foundation for Strategic Research (SSF), and European Union (EU) Interreg Program. The computations and data handling were enabled by resources provided by the Swedish National Infrastructure for Computing (SNIC) partially funded by the Swedish Research Council (VR) through grant agreement no. 2018-05973.}

\author[1]{Amin Aminifar}[
                        orcid=0000-0002-9920-2539]

\cormark[1]

\ead{amin.aminifar@ziti.uni-heidelberg.de}

\affiliation[1]{organization={Institute of Computer Engineering, Heidelberg University},
    city={Heidelberg},
    postcode={69120}, 
    country={Germany}}

\author[2]{Matin Shokri}[orcid=0000-0002-2992-9329]
\cormark[1]

\affiliation[2]{organization={Faculty of Computer Engineering, K. N. Toosi University of Technology},
    city={Tehran},
    postcode={19697-64499}, 
    country={Iran}}
\ead{shokri@email.kntu.ac.ir}

\author[3]{Amir Aminifar}[%
   orcid=0000-0002-1673-4733
   ]

\ead{amir.aminifar@eit.lth.se}


\affiliation[3]{organization={Department of Electrical and Information Technology, Lund University},
    city={Lund},
    postcode={22100}, 
    country={Sweden}}

\newcommand{\hl}[1]{\color{blue} #1 \color{black}}

\newcommand\blfootnote[1]{%
  \begingroup
  \renewcommand\thefootnote{}\footnote{#1}%
  \addtocounter{footnote}{-1}%
  \endgroup
}
\cortext[1]{Equal contribution.}

\begin{abstract}
Machine Learning (\gls{ml}) algorithms are generally designed for scenarios in which all data is stored in one data center, where the training is performed. 
However, in many applications, e.g., in the healthcare domain, the training data is distributed among several entities, e.g., different hospitals or patients' mobile devices/sensors. At the same time, transferring the data to a central location for learning is certainly not an option, due to privacy concerns and legal issues, and in certain cases, because of the communication and computation overheads.  
\gls{fl} is the state-of-the-art collaborative \gls{ml} approach for training an \gls{ml} model across multiple parties holding local data samples, without sharing them. 
However, enabling learning from distributed data over such edge \gls{iot} systems (e.g., mobile-health and wearable technologies, involving sensitive personal/medical data) in a privacy-preserving fashion presents a major challenge mainly due to their stringent resource constraints, i.e., limited computing capacity, communication bandwidth, memory storage, and battery lifetime. In this paper, we propose a privacy-preserving edge \gls{fl} framework for resource-constrained mobile-health and wearable technologies over the \gls{iot} infrastructure. We evaluate our proposed framework extensively and provide the implementation of our technique on Amazon's AWS cloud platform based on the seizure detection application in epilepsy monitoring using wearable technologies. 
\end{abstract}




\maketitle

\section{Introduction}
\label{sec:intro}
Over the past few decades, we have been witnessing a dramatic increase in healthcare costs. In the United States of America, for instance, the healthcare costs have risen from 27.2 billion dollars (only 5 percent of GDP) in 1960 to 41.2 billion dollars (19.7 percent of GDP) in 2020 \cite{NHEA2020,hartman2022national}. Current practice in the healthcare system places a heavy burden on the society, in terms of healthcare expenditures, due to its treatment-oriented philosophy. Mobile-health and wearable technologies offer a promising solution to pervasive healthcare by removing the constraints with respect to time and location \cite{Varshney07MNA}. \blfootnote{Published in Future Generation Computer Systems (\nolinkurl{https://doi.org/10.1016/j.future.2024.07.035})}

Several real-time mobile-health technologies for monitoring critical health conditions have been proposed in the past few years \cite{Surrel18TBIOCAS,Sopic17BIOCAS,Sopic18TBIOCAS, Sopic18ISCAS,rahmani2018exploiting,Forooghifar2019MONET}. The prime examples include real-time epilepsy monitoring and seizure detection for epilepsy patients \cite{Sopic18ISCAS,Forooghifar2019MONET} and real-time cardiac monitoring and heart-attack detection for patients prone to heart attack \cite{Sopic17BIOCAS,Sopic18TBIOCAS}, to notify the family members, caregivers, and emergency units for rescue in the case of such life-threatening events. As a result, mobile-health and wearable technologies, have the potential to not only improve the quality of life for these patients, but also reduce the mortality rate associated with such health conditions.

The majority of the state-of-the-art detection techniques for such adverse health events, i.e., epileptic seizures, strokes, heart attacks, adopt modern \gls{ml} algorithms, e.g., deep neural networks \cite{hannun2019cardiologist, ribeiro2020automatic, baghersalimi2021personalized, amirshahi2022m2d2}. However, training reliable deep neural network models requires a large amount of data, i.e., the so-called Big Data. As a result, the training is often performed based on the data collected from many patients, which are stored and distributed on patients' mobile devices and wearable sensors. Traditionally, all training data samples are assumed to be present where the \gls{ml} algorithm runs, i.e., the assumption was that the learning algorithm has access to all training data samples throughout the entire learning process. Therefore, to learn a model by adopting conventional solutions, each party holding a portion of the data is required to transfer all its training data in raw format to a center for the learning process. However, such an approach raises major privacy concerns in connection to sharing sensitive personal/patient data \cite{lustgarten2020digital,Pascual2020epilepsygan}.

Here, we address the problem of learning from decentralized healthcare data on mobile-health and wearable technologies, while preserving the patients' privacy. We focus on real-time epileptic seizure detection application using mobile-health and wearable technologies. Epilepsy is one of the most common neurological diseases affecting more than 50 million people worldwide~\cite{epilepsy} and is ranked number four after migraine, Alzheimer's disease, and stroke~\cite{Hirtz}, with an economic burden estimated at 15.5 billion euros per year in Europe \cite{Pugliatti07Epilepsia}. Despite the recent advances in anti-epileptic drugs, one-third of the epilepsy patients still suffer from seizure. More importantly, epilepsy represents the second neurological cause of years of potential life lost, primarily due to seizure-triggered accidents and sudden unexpected death in epilepsy (SUDEP)~\cite{thurman2014}. The possibility of monitoring epileptic patients in real time and on a long-term basis is likely to improve the quality of life and reduce the mortality rate in these patients. In particular, based on such real-time epileptic seizure detection mechanisms, it is possible to notify the family members, caregivers, and emergency units in case of a seizure for rescue. This will provide the opportunity of reducing the mortality rate due to seizure-related injuries, status epilepticus, and SUDEP~\cite{Lancet}.

In this article, we assume the data is horizontally partitioned, i.e., all attributes of each data sample are stored together on one party. We propose a privacy-preserving \gls{fl} framework for mobile-health and wearable technologies. Our proposed framework is specifically designed to consider the extreme resource constraints of mobile-health and wearable technologies, i.e., in terms of computing power, communication bandwidth, and battery lifetime. \color{black}This is ensured by leveraging on the already-collected high-quality data in the hospital environments and then partially refining the distributed deep neural network model based on the data acquired by mobile-health and wearable systems, to reduce the computing and communication overheads of \gls{fl} for such resource-constrained devices, while maintaining the prediction performance.\color{black}

From a privacy perspective, our proposed framework ensures that the local training data does not leave patients' devices by adopting \gls{fl} scheme. Moreover, our framework guarantees that the partial contribution of each patient during the learning process remains private, by adopting \gls{smc} techniques. In particular, we show that our privacy-preserving \gls{fl} framework is secure even under the collusion of $k$ parties involved in the learning process. Finally, we present the implementation of our framework on Amazon's AWS cloud platform and evaluate our technique based on the CHB-MIT dataset \cite{goldberger2000physiobank,shoeb2009application} and smart mobile-health and wearable applications \cite{Sopic18ISCAS}.

In this work, we aim at developing a privacy-preserving edge \gls{fl} framework for mobile-health and wearable technologies with stringent resource constraints. Our main contributions are summarized below:
\begin{itemize}
    \item a privacy-preserving edge \gls{fl} framework for resource-constrained mobile-health and wearable technologies over the \gls{iot} infrastructure; and 
    \item evaluation and implementation of our proposed framework, including on Amazon's AWS cloud, based on real-world \gls{iot} applications in the healthcare domain.
\end{itemize}

The paper is structured as follows. Section \ref{sec:StateOfTheArt} reviews the state of the art in \gls{fl}, including in the context of the \gls{iot} systems and health applications, with a main focus on privacy. Section \ref{sec:Method} illustrates the proposed framework for privacy-preserving edge \gls{fl}, designed for resource-constrained mobile-health and wearable technologies. Section \ref{sec:Evaluation} evaluates the proposed framework on real-world \gls{iot} applications, including implementation of our proposed framework on Amazon's AWS cloud. Finally, Section \ref{sec:Conclusion} serves as the conclusion of this article.

\section{State of the Art}
\label{sec:StateOfTheArt}

The naive solution to learning from decentralized data is to share the data with a trusted party, but this is not feasible in many applications due to privacy and/or legal concerns \cite{emekcci2007privacy}. One solution for addressing such privacy concerns is to publish an altered copy of data by employing privacy-preserving data publishing methods \cite{fung2010introduction,dwork2006differential,pascual2019synthetic}, e.g., anonymization methods \cite{sweeney2002k,lefevre2006mondrian,machanavajjhala2007diversity,aminifar2021diversity}. However, the conventional privacy-preserving data publishing methods are also not designed for scenarios in which the data is distributed. Several studies propose methods for collaborative anonymization when data is distributed over multiple data holders \cite{jurczyk2009distributed, mohammed2010centralized,luo2013distributed}. Nevertheless, such methods generally suffer from the degradation of data utility in exchange for protecting the privacy of data subjects, i.e., the trade-off between privacy and utility \cite{apple2017}.

\gls{fl}, introduced in \cite{konevcny2016federated}, is a powerful solution for training \gls{ml} models in scenarios in which the training data is decentralized. \gls{fl} keeps the training data distributed among parties/clients, e.g., patients' personal mobile devices, and learns a model collaboratively from such decentralized data, usually with a central server for the orchestration of the process. This approach follows the data minimization principle in data protection guidelines \cite{house2012consumer,GDPR_principles} and limits the privacy risks that we face in centralized \gls{ml} \cite{mcmahan2016communication}. 

In the past few years, a large number of studies consider \gls{fl} to address their concerns related to privacy and overhead, particularly in the healthcare domain \cite{yang2019federated,silva2019federated,baghersalimi2021personalized}. 
\gls{fl} methods are usually based on neural network algorithms, but the term has also been used for other \gls{ml} algorithms recently, e.g., for tree-based algorithms \cite{truex2019hybrid, liu2020federated}, for algorithms that are both gradient-based and tree-based \cite{cheng2021secureboost, liu2020boosting, zhao2018inprivate, li2020practical, li2023fedtree}, or for other algorithms \cite{polato2022boosting,mittone2023model}.
Several literature reviews of the state-of-the-art \gls{fl} techniques are available in \cite{kairouz2019advances,li2021survey,lo2021systematic}.

\color{black}
Several studies focus on tree-based algorithms to train a model based on decentralized data. For instance, in \cite{lindell2002privacy, emekcci2007privacy, liu2020federated, aminifar2021privacy, aminifar2021Scalable,aminifar2022extremely} the ID3 \cite{quinlan1986induction}, random forest \cite{ho1995random}, and extremely randomized trees \cite{geurts2006extremely} algorithms are extended to be utilized for scenarios in which the training data is decentralized. These methods employ encryption or SMC to protect the privacy of data holders. Several studies concentrate on the AdaBoost algorithm \cite{freund1997decision}, which is commonly integrated with tree-based techniques \cite{zharmagambetov2021improved}, and extend it for distributed scenarios \cite{lazarevic2002boosting, cooper2017improved, polato2022boosting}. A notable exception is the study by Mittone et al. \cite{mittone2023model}, which proposes the Model-Agnostic Federated Learning (MAFL) system that merges AdaBoost with Intel OpenFL \cite{foley2022openfl} that is not tied to any machine learning model. While tree-based methods can outperform standard deep neural networks when dealing with structured or tabular data where features are individually meaningful \cite{lundberg2020local}, they generally have an inferior performance compared to deep neural networks for data with strong multi-scale temporal or spatial structures. 
\color{black}

\gls{ml} algorithms, when employed in the context of IoT devices and wearable systems, should be adapted to consider the resource constraints in such systems \cite{ragusa2019survey,ragusa2021hardware,pandelea2021emotion,liu2022lgc,ragusa2023random}. In particular, in cross-device \gls{fl}, where we have a large number of mobile or IoT devices, communication is often the main bottleneck \cite{kairouz2019advances}. In \cite{mcmahan2016communication}, the authors propose FederatedAveraging (FedAvg) in order to reduce the rounds of communication required for training a deep neural network, by selecting only a fraction of clients and by introducing local updates. In \cite{li2020federated}, the authors propose FedProx, which is a generalization of FedAvg, basically allowing variable amounts of work to be performed on client devices depending on the available resources. 
In \cite{wang2019adaptive}, the authors address the problem of training gradient-descent-based machine-learning models with data distributed on edge devices. This study makes a tradeoff between local updates on clients and global updates on the central server to minimize the loss function with a given resource budget. In \cite{nishio2019client}, the authors propose FedCS to address the challenges related to efficiency in the application of \gls{fl} for clients with limited resources, by managing the clients based on the current condition of their resources. However, privacy preservation has not yet been explored in the context of \gls{fl} on IoT platforms and in the context of mobile-health applications.

As discussed earlier, data privacy in the context of medical applications is absolutely essential \cite{castiglione2013secure,Pascual2020epilepsygan,kaur2022trustworthy}. 
During the past few years, there has been a large number of studies discussing the privacy risks associated with information leakage through sharing gradients \cite{shokri2015privacy,bonawitz2016practical,aono2017privacy,hitaj2017deep}. To this end and to address recent attacks on the neural networks, e.g., membership inference attack \cite{nasr2019comprehensive, truex2019demystifying}, several studies adopt differential privacy \cite{dwork2006differential} in their methods \cite{wei2020federated, hu2020personalized, geyer2017differentially}. For instance, in \cite{wei2020federated}, the authors combine \gls{fl} with differential privacy. In essence, the authors propose to add artificial noises to the parameters on the clients' side before model aggregation to provide differential privacy. In \cite{hu2020personalized}, the authors propose an \gls{fl} scheme with differential privacy guarantee and evaluate their scheme based on realistic mobile sensing data. In \cite{geyer2017differentially}, the authors present a federated optimization algorithm by considering differential privacy to hide clients' contributions in the process of learning. The differential privacy framework, however, makes a trade-off between privacy and data utility, hence generally leading to a degradation of the final prediction performance \cite{davis2019improving}.

A comprehensive study on the security and privacy of \gls{fl} is performed in \cite{mothukuri2021survey}. The authors extensively discuss the vulnerabilities and threats in \gls{fl}. For instance, one concern in \gls{fl} is data poisoning \cite{biggio2012poisoning}, which is relevant when the attacker incorporates malicious data points to training data in order to maximize the error. 
Membership inference attack \cite{nasr2019comprehensive, truex2019demystifying} is another concern that enables determining whether a specific record was included in the data used to train a model.

Several state-of-the-art studies propose solutions for learning from decentralized data that are based on homomorphic cryptography or \gls{smc} \cite{bost2015machine,graepel2012ml,bos2014private,gilad2016cryptonets}. A comprehensive review of studies for such privacy-preserving data mining approaches for horizontally and vertically partitioned data is available in \cite{kantarcioglu2008survey,vaidya2008survey}. 
One of the main challenges with homomorphic cryptography and \gls{smc} is their major communication and computation overhead, which makes them impractical in many real-world scenarios with IoT and mobile-health applications \cite{emekcci2007privacy,pinkas2002cryptographic,vaidya2013random}. Despite several attempts to improve the computational complexity of homomorphic encryption, including somewhat homomorphic encryption and learning-with-error, the existing homomorphic encryption schemes still suffer from extreme computational complexity \cite{naehrig2011can}. 
Therefore, new solutions are required to also consider the overheads related to communication and computation, which are particularly relevant in the context of resource-constrained mobile-health and wearable systems \cite{zhang2021current}.

\color{black}Finally, several studies have considered \gls{fl} in the context of medical applications. For instance, Wu et al. \cite{wu2020fedhome} adopt a cloud-edge personalized \gls{fl} for in-home health monitoring for fall detection, based on 57 gyroscope and accelerometer recordings collected using smartphones. Similarly, Yoo et al. \cite{yoo2021personalized} use \gls{fl} for multiple patients to detect Major Depressive Disorder based on cardiac function. Yang et al. \cite{yang2022hypernetwork} adopt \gls{fl} for simulated multi-institutional Computed Tomography (CT) imaging. Chen et al. \cite{chen2022personalized} use \gls{fl} in the context of multi-institutional Prostate MRI and Dermoscopic datasets. In our previous work \cite{baghersalimi2021personalized,baghersalimi2023decentralized}, we have proposed \gls{fl} schemes for seizure detection. None of these previous studies, however, take into consideration the privacy concerns.\color{black}

In our previous studies in \cite{Sopic18ISCAS,Surrel18TBIOCAS,Sopic18TBIOCAS,Forooghifar2019MONET,de2020real,zanetti2021real,forooghifar2021self}, we have developed energy-efficient \gls{ml} techniques for mobile-health and wearable technologies, including in the distributed and federated learning \cite{forooghifar2019resource,baghersalimi2021personalized, baghersalimi2023decentralized}. On the other hand, we have looked into privacy-preserving distributed and federated learning techniques considering tree-based algorithms \cite{aminifar2021Scalable, aminifar2021privacy, aminifar2022extremely, aminifar2022privacy}, but not considering the resource-constraints of mobile-health and wearable technologies. To address this gap, in this article, we propose a framework that jointly considers prediction performance, computation and communication overheads, and privacy concerns, which are all essential for resource-constrained mobile-health systems involving sensitive medical/personal data.

\section{Privacy-Preserving \gls{fl}}
\label{sec:Method}
In this section, we describe our proposed privacy-preserving \gls{fl} framework for mobile-health and wearable technologies. Section \ref{subsec:privacy_model} will briefly discuss the privacy model we consider in this paper. Then, in Section \ref{subsec:federated_learning}, we discuss our proposed privacy-preserving edge \gls{fl} framework for mobile-health and wearable technologies.

\subsection{Privacy Model}\label{subsec:privacy_model}
Contrary to popular belief, the security and privacy of non-invasive mobile health technologies are of critical importance, even though there is no explicit close-loop intervention involved \cite{pascual2019synthetic, Pascual2020epilepsygan}. In this article, we focus only on those attacks targeting the privacy of the patients. The privacy of the patients can be compromised if the information communicated between wearable sensors, mobile phones, or clouds is disclosed to unauthorized entities. This becomes all the more critical if the information communicated includes medical records of the patients. 

The model in this article considers the privacy among the parties involved in the computation. We consider the following assumptions in this privacy model:
\begin{itemize}
\item We assume the \emph{honest-but-curious} model for all parties involved in the collaborative computation, including the local sensors and cloud engines. That is, the parties involved in the collaborative computation do not actively interfere with the computation, i.e., neither provide incorrect information, nor manipulate the information they receive from other parties involved in the computation. 

\item Moreover, we assume that the number of colluding parties is less than $k$. 
The parameter $k$ can be adapted considering the sensitivity of the information shared within the learning process, i.e., the higher the value of $k$, the more parties need to collude to reveal secret values. 
\end{itemize}
\color{black}
\subsection{Privacy-Preserving Edge \gls{fl}} \label{subsec:federated_learning}
We shall now present our proposed privacy-preserving edge \gls{fl} in two steps. In the first step, in Section \ref{subsec:Initialization}, we illustrate the initialization procedure among healthcare centers to train an initial base neural network model in a privacy-preserving fashion. Then, in Section \ref{subsec:Procedure}, we discuss our privacy-preserving \gls{fl} framework for mobile-health and wearable technologies, which reduces the complexity by exploiting the the initial base neural network model trained in Section \ref{subsec:Initialization}.

\subsubsection{\gls{fl} Base Model}
\label{subsec:Initialization}
\begin{figure}
        \centering
        \includegraphics[width=\columnwidth]{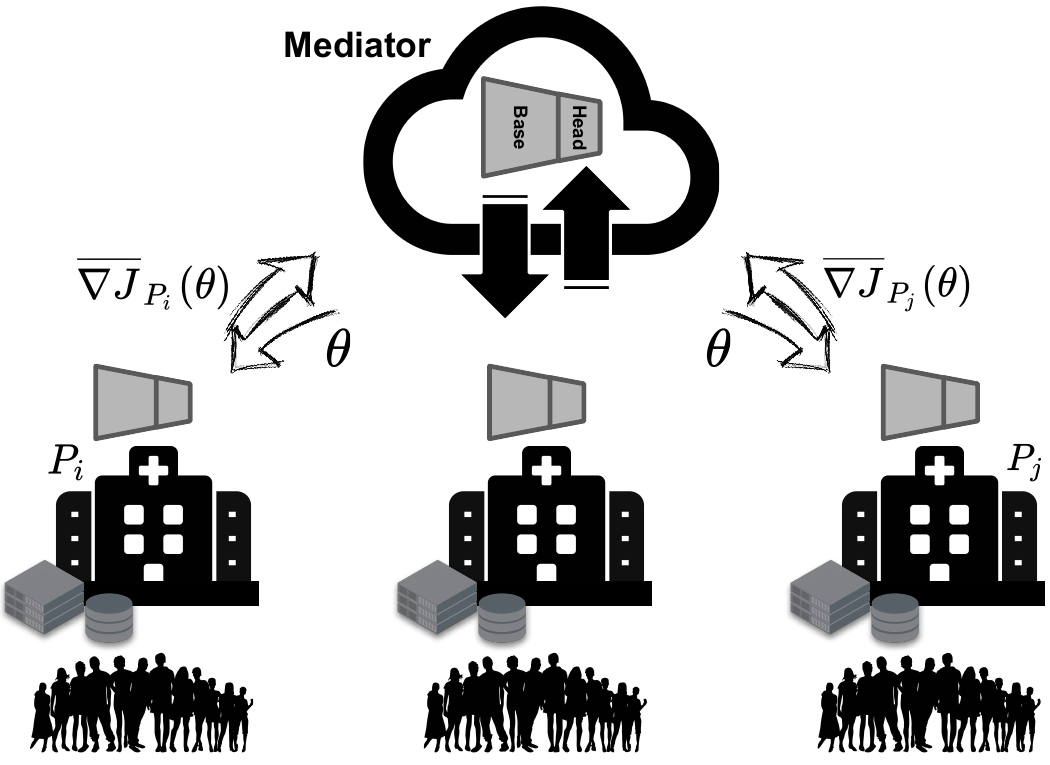} 
        \caption[caption]{Overview of the initialization phase in our proposed \gls{fl} framework} 
        \label{fig:initial_phase}
\end{figure}

\color{black}
In the initialization phase, we assume $N$ hospitals or healthcare centers are involved in training an initial generic model for epileptic seizure detection. The training data is distributed among all healthcare centers involved and the training is required to be performed in a secure and distributed fashion, since the personal/patient medical data is generally considered sensitive and may not leave hospital premises due to regulatory restrictions. This scenario is shown in Figure \ref{fig:initial_phase}.

In the first step, all healthcare centers will exchange their private keys pair-wise in a secure fashion, using the Diffie-Hellman \cite{Diffie1976New} key exchange scheme. In other words, each two healthcare centers will have a shared key in a secure way. The shared keys are used in the secure \gls{fl} scheme explained below, where we assume that the healthcare centers have sufficient computing, communication, and storage resources.

Given a deep neural network model and the data, the goal is to train a robust and reliable model based on the training data available on all healthcare centers. The model (hypothesis) is shown by $h_{\theta}(X)$, in which $\theta$ is the set of the network's parameters that are optimized during the process of learning, and $X$ is the input to the network. The objective is to minimize the loss represented by the following function:
\noindent
{\begin{align}\label{eq:cost}
J(\theta) = \frac{1}{2m} \sum_{i=1}^{m}( h_{\theta}(X^{(i)})-y^{(i)})^{2},
\end{align}}

\noindent
where $m$ represents the number of training data samples. $X^{(i)}$ denotes the attributes for the $i^\textrm{th}$ training data sample, and $y^{(i)}$ is the target value for the $i^\textrm{th}$ training data sample.
Having the loss function $J(\theta)$, the model parameters $\theta$, the training data $X$, and the labels $y$, we compute the gradients for updating the model parameters. 
Having the gradients, we update the model parameters based on an optimization algorithm, e.g., the Adam optimizer. This process summarizes the conventional centralized training procedure for DNNs. 

\gls{fl} is the state-of-the-art technique for learning from decentralized data. 
In such settings, since all data samples $X$ and labels $y$ are not stored in one data center, we calculate the partial derivatives in a way that the summation of them, from all parties, is equal to the gradient of all training data samples, which we refer to it as the global gradient. 
In a decentralized scenario, the training data samples are distributed among $n$ parties. In general, we have $m$ training data samples, and Party $j$ holds $m_j$ data samples of the total $m = \sum_{j=1}^{n} m_{j}$ samples.
The partial gradients are obtained by calculating the gradient locally at each party, i.e., considering the $m_j$ available data sample at Party $P_j$. By aggregating the partial gradients received from all parties, we calculate the global gradient. The following shows how we calculate the partial gradient at Party $j$ ($P_j$):
\begin{align}\label{eq:grad}
\hspace{-0.7cm}\nabla J_{P_j}(\theta) = 
\sum_{i=1}^{m_j} (h_{\theta}(X_{P_j}^{(i)})-y_{P_j}^{(i)}) \frac{\partial h_{\theta}(X_{P_j}^{(i)})}{\partial \theta}.
\end{align}

The standard \gls{fl} scheme, however, while reducing the privacy concerns, does not guarantee privacy. Therefore, the data-holder parties may not directly share their contribution, partial gradients, with other parties for privacy and legal concerns \cite{shokri2015privacy,bonawitz2016practical,aono2017privacy,hitaj2017deep}. Several studies incorporate noise to data-holder parties' contribution in order to provide differential privacy \cite{geyer2017differentially}. However, this degrades the prediction performance as there is a trade-off between privacy and data utility in differential privacy \cite{davis2019improving}. In contrast, here, the optimizer merely requires the global gradient to update the network parameters and we do not require parties to share their partial gradients in their raw format. Therefore, in this study, for calculating the global gradient, we employ a secure aggregation technique. In essence, our proposed framework masks the contributions of each data-holder party to preserve privacy. Towards this, each party includes its local masks to the partial gradients and only then shares the results with the mediator (a central server orchestrating the training process):

\begin{align}\label{eq:masked_grad}
\overline{\nabla J}_{P_j}(\theta) = \nabla J_{P_j}(\theta)  + \sum_{i=1}^{n} R_{j,i} - \sum_{i=1}^{n} R_{i,j},
\end{align}
where $R_{i,j}$ and $R_{j,i}$ are the masks shared between data-holder party $j$ and data-holder party $i$. The masks $R_{i,j}$ and $R_{j,i}$ between data-holder party $j$ and data-holder party $i$ may be arbitrary values obtained based on the private key shared between party $i$ and party $j$. The masks $R_{i,j}$ are to ensure privacy of the secret value of party $i$, while the masks $R_{j,i}$ are to ensure privacy of the secret value of party $j$. In essence, the proposed protocol ensures that each mask is added exactly once and subtracted exactly once in careful coordination, such that in the final aggregation results all masks are canceled out. For the simplicity of the presentation, we will drop the argument $\theta$ in $\overline{\nabla J}_{P_j}(\theta)$ and $\nabla J_{P_j}(\theta)$, denote them by $\overline{\nabla J}_{P_j}$ and $\nabla J_{P_j}$, respectively.

In the next step, the mediator aggregates all masked partial gradients as follows: 
\color{black}
\begin{align}\label{equ:masked_aggregation_1}
\hspace{-0.7cm}\frac{1}{m} \cdot \sum_{j=1}^{n}  \overline{\nabla J}_{P_j} &= \frac 1 m \sum_{j=1}^{n} \left( \nabla J_{P_j}  + \sum_{i=1}^{n} R_{j,i} - \sum_{i=1}^{n} R_{i,j} \right)\\\nonumber
&= \frac 1 m \sum_{j=1}^{n} \nabla J_{P_j}+ \underbrace{\frac 1 m  \sum_{j=1}^{n} \left( \sum_{i=1}^{n} R_{j,i} - \sum_{i=1}^{n} R_{i,j} \right)}_{\textrm{overall mask = 0}},\nonumber
\end{align}
where the value of the second part, i.e., \emph{overall mask}, is zero, because $\sum_{j=1}^{n} \sum_{i=1}^{n} R_{j,i} = \sum_{j=1}^{n} \sum_{i=1}^{n} R_{i,j}$. Next, we aggregate the value of $\nabla J_{P_j}$ for all parties, as follows:
\begin{align}\label{equ:masked_aggregation_2}
\hspace{-0.7cm}\frac{1}{m} \cdot \sum_{j=1}^{n}  \overline{\nabla J}_{P_j} &= \frac 1 m \sum_{j=1}^{n}  \nabla J_{P_j} \\\nonumber
&= \frac{1}{m} \cdot \sum_{j=1}^{n}  \sum_{i=1}^{m_j} (h_{\theta}(X_{P_j}^{(i)})-y_{P_j}^{(i)}) \frac{\partial}{\partial \theta} h_{\theta}(X_{P_j}^{(i)})\\\nonumber
                 &= \frac{1}{m} \cdot \sum_{i=1}^{m}  (h_{\theta}(X^{(i)})-y^{(i)}) \frac{\partial}{\partial \theta} h_{\theta}(X^{(i)})\\\nonumber
&= \nabla J.
\end{align}
\color{black}
The above calculations show that the aggregated partial gradients over all involved parties will provide the global gradients in a secure and privacy-preserving fashion.
The global gradient, i.e., $\nabla J$, will then be used to update the deep neural network parameters.

In summary, in this initialization stage, the following procedure is performed iteratively: (1) all healthcare centers exchange pair-wise keys for privacy-preserving \gls{fl}, (2) all data-holder parties calculate the masked partial gradients, i.e., $\overline{\nabla J}_{P_j}$ for party $j$ (Equation \ref{eq:masked_grad}), and share them with the mediator, (3) the mediator aggregates the masked partial gradients and obtains the global gradient, i.e., $\nabla J$, and updates the deep neural network parameters accordingly, and (4) the mediator shares the updated deep neural network model with data-holder parties.

The final outcome in this section is an initial generic neural network model trained collaboratively among the healthcare centers in a privacy-preserving fashion. While this procedure can be performed by the healthcare centers, which are not limited in terms of computing, communication, and storage resources, the same procedure is infeasible for mobile-health and wearable technologies with extreme resource constraints. At the same time, the initial generic neural network modeled in this procedure is specifically trained for high-quality data acquired by high-tech hospital equipment. 
Therefore, the model trained here on high-quality data from the high-tech hospital equipment often performs poorly on the data acquired by the mobile-health and IoT devices during ambulatory monitoring.
In the next section, we propose a privacy-preserving edge \gls{fl} scheme to exploit the initial generic neural network model trained in this section for training deep neural networks on resource-constrained mobile-health and IoT technologies.

\color{black}
\subsubsection{\gls{fl} on Edge}
\label{subsec:Procedure}
\begin{figure}
        \centering
        \includegraphics[width=\columnwidth]{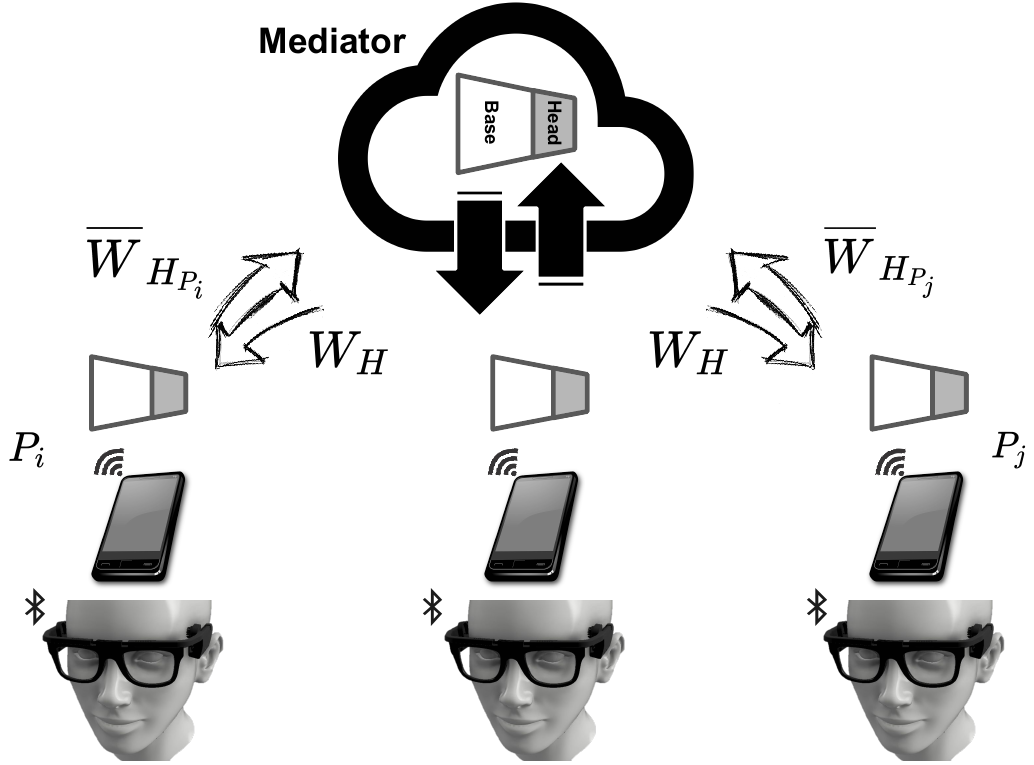}   
        \caption[ caption ]
        {Overview of our proposed edge \gls{fl} framework on resource-constrained mobile-health devices} 
        \label{fig:federated_learning}
\end{figure}

\color{black}
In this section, we discuss our proposed privacy-preserving \gls{fl} framework for mobile-health and wearable technologies. The overview of our proposed privacy-preserving \gls{fl} is shown in Figure \ref{fig:federated_learning}.

In the first step, similar to the previous section, the key exchange protocol via the Diffie-Hellman scheme \cite{Diffie1976New} is adopted to exchange secret shares among the parties in a pair-wise fashion. The Diffie-Hellman key exchange scheme is, however, relatively complex when it comes to mobile-health and wearable technologies. To alleviate this, instead of executing the Diffie-Hellman key exchange scheme in every iteration of the learning process, we propose to share a secret random seed pair-wise among the parties in the initialization phase. This seed will then be used to generate secret masks during the entire process of privacy-preserving \gls{fl}.

During the \gls{fl} process, each patient updates its model based on the local data, e.g., based on stochastic/batch gradient decent, as follows:
\begin{align}
W_{P_j}^{(s)} &= W_{P_j}^{(s-1)}- \alpha \cdot \nabla J^{(s-1)}_{P_j},
\end{align}
where $W_{P_j}^{(s)}$ captures the neural network model/weights for the patient $P_j$, in iteration $s$. The coefficient $\alpha$ is used to adapt the convergence rate during the gradient descent algorithm. 

The key challenge here is that training this model from scratch incurs major overheads for resource-constrained mobile-health and wearable technologies. On the one hand, while the initial generic neural network model trained collaboratively by the healthcare centers (in Section \ref{subsec:Initialization}) is tailored to the high-quality data from the high-tech hospital equipment, this model can still be used as a basis for the data acquired by the mobile-health and wearable technologies. This will allow faster convergence of the model, hence reducing the communication and computation overheads of the training. 

Unfortunately, training the entire neural network, even starting from the already trained models in the initial step, is still computationally infeasible for the resource-constrained mobile-health and wearable technologies. To address this challenge, inspired by the transfer learning techniques, we propose to only refine the model partially, where only the last few layers are fine-tuned to the data acquired by the mobile-health and wearable devices. More specifically, the initial generic model, denoted by $M$, is already available. As shown in Figure \ref{fig:federated_learning}, the model $M=[B\rightarrow H]$ consists of two parts: the base architecture, denoted by $B$, and the head architecture, denoted by $H$. To reduce the overheads of training on resource-constrained mobile-health and wearable devices, we only train the head architecture $H$ on the mobile-health devices in a federated and privacy-preserving fashion. The base architecture $B$, while trained on high-quality data from high-tech hospital equipment, is still able to capture patterns which are also relevant in the case of the data acquired by the mobile-health and wearable devices. Mathematically, this is captured by the following equation:
\begin{align}
W_{H_{P_j}}^{(s)} &= W_{H_{P_j}}^{(s-1)}- \alpha \cdot \nabla J^{(s-1)}_{H_{P_j}},
\end{align}
where $W_{H_{P_j}}^{(s)}$ captures the neural network model/weights of the head architecture $H$ for the patient $P_j$, in iteration $s$. 
\color{black}There is an inherent trade-off between the number of layers that are to be fine-tuned and the prediction performance of the \gls{fl} model. That is, the more the number of the layers we fine-tune, the higher the prediction performance and the computation overheads. One strategy to strike a balance between the performance and complexity is to iterate over all layers and select the base--head partitioning with the best complexity--performance trade off based on the validation data. This optimization has a linear time-complexity and will be performed at design time, hence tractable even in deep neural networks. In practice and as shown in our experimental results, however, refining a small proportion of the parameters, e.g., often only the last layer, is sufficient to obtain on par performance with training the entire neural network.\color{black}

The process of \gls{fl} on mobile-health and wearable devices, however, still requires the base architecture $B$. While \gls{fl} is only targeted at the head architecture $H$, the feed-forward through the base architecture $B$ is unavoidable (note that the back-propagation is not required since the base architecture is not to be retrained). To address this challenge, the mediator provides a distilled base architecture $\hat{B}$, which dramatically reduces the complexity of the training the head architecture $H$ \cite{hinton2015distilling}. Therefore, our new model is captured by $\hat{M}=[\hat{B}\rightarrow H]$, where we freeze the distilled base architecture $\hat{B}$ to reduce the overheads.

Then, each party will share the masked final deep neural network parameters of the head architecture $H$ with the mediator as follows,
\begin{align}
\hspace{-0.7cm}\overline{W}_{H_{P_j}}^{(s)} &= W_{H_{P_j}}^{(s-1)}- \alpha \cdot \nabla J^{(s-1)}_{H_{P_j}} + \sum_{i=1}^{n} R_{j,i}^{(s-1)} - \sum_{i=1}^{n} R_{i,j}^{(s-1)},
\end{align}
where $\overline{W}_{H_{P_j}}^{(s)}$ captures the neural network model/weights of the head architecture $H$ for the patient $P_j$, in iteration $s$. The above secret sharing scheme, however, requires a pair-wise Diffie-Hellman key exchange, which incurs major overheads for the \gls{iot} and mobile-health devices. Therefore, here, we proposed a new scheme where each patient $P_j$ is only required to have Diffie-Hellman key exchange with only $k$ other patients. \color{black}These $k$ patients/parties may be selected arbitrarily. In the presence of $k$ trustworthy/reliable patients/parties, each patient may select their $k$ trustworthy/reliable patients/patients independently; otherwise, the $k$ patients/patients may be selected arbitrarily/randomly. \color{black} The set of patients with whom Patient $P_j$ will exchange keys is denoted by $S_j$. Note that if $i\in S_j$, then we are sure that $j\in S_i$, hence $R_{i,j}=R_{j,i}$. The above equation then can be reformulated considering this assumption as follows, 
\begin{align}
\hspace{-0.7cm}\overline{W}_{H_{P_j}}^{(s)}\! &=\! W_{H_{P_j}}^{(s-1)} \!-\! \alpha \cdot \nabla J^{(s-1)}_{H_{P_j}} \! + \! \sum_{\forall i\in S_j} R_{j,i}^{(s-1)}  \! -\! \sum_{\forall j \in S_i} R_{i,j}^{(s-1)},
\end{align}
where the design parameter $k$, that is the size of set $S_j$ and $S_i$, can be adjusted to make a trade-off between the computational/communication overheads and privacy/security strength.

Finally, the mediator will aggregate all deep neural network parameters associated with the head architecture to obtain the global deep neural network model of the head architecture as follows,

\begin{align}
\hspace{-0.7cm} \frac{1}{n} \sum_{j=1}^{n} \overline{W}_{H_{P_j}}^{(s)}
&= \frac{1}{n} \sum_{j=1}^{n}\biggl( W_{H_{P_j}}^{(s-1)}- \alpha \cdot \nabla J^{(s-1)}_{H_{P_j}}\\\nonumber
&+ \sum_{\forall i\in S_j} R_{j,i}^{(s-1)} - \sum_{\forall j \in S_i} R_{i,j}^{(s-1)} \biggr)\\\nonumber
&= \frac{1}{n} \sum_{j=1}^{n}\left( W_{H_{P_j}}^{(s-1)}- \alpha \cdot \nabla J^{(s-1)}_{H_{P_j}} \right)\\\nonumber
&+ \frac{1}{n} \sum_{j=1}^{n}\biggl(\sum_{\forall i\in S_j} R_{j,i}^{(s-1)} - \sum_{\forall j \in S_i} R_{i,j}^{(s-1)} \biggr)\\\nonumber
&= \frac{1}{n} \sum_{j=1}^{n}\left( W_{H_{P_j}}^{(s-1)}- \alpha \cdot \nabla J^{(s-1)}_{H_{P_j}} \right)\\\nonumber
&= W_{H}^{(s-1)}- \alpha \cdot \nabla J^{(s-1)}_{H} \\\nonumber
&=W_{H}^{(s)},
\end{align}
where the global deep neural network parameters associated with the head architecture in iteration $s$ is denoted by $W_{H}^{(s)}$. In the third equality, we make use of the following equality, 
\begin{align}
\sum_{j=1}^{n}\biggl(\sum_{\forall i\in S_j} R_{j,i}^{(s-1)} - \sum_{\forall j \in S_i} R_{i,j}^{(s-1)} \biggr)=0,
\end{align}
which holds because the pair-wise random masks are once added by one party in each pair and once subtracted by the other party in each pair. 

\color{black}
Our proposed framework also offers the possibility of personalization, to enable the realization of the precision medicine paradigm. Once the \gls{fl} process discussed above is completed, the head $H$ can be immediately refined based on the data locally available to each patient. In this way, the parameters in the head $H$ architecture will be fine-tuned to the specific profile of each patient, enabling precision detection of health pathology, as we show experimentally in the next section. 
\color{black}

In summary, our proposed privacy-preserving edge \gls{fl} framework is specifically designed for resource-constrained \gls{iot}, mobile-health, and wearable technologies, not only from the \gls{fl} perspective, but also from a privacy perspective. In the next section, we will evaluate our proposed privacy-preserving edge \gls{fl} framework with respect to prediction performance, privacy, and overheads.

\section{Experimental Setup and Results}
\label{sec:Evaluation}
This section presents the experimental results and evaluation of our proposed framework with respect to classification performance, overhead, and privacy, which are the main criteria for the assessment of privacy-preserving data mining algorithms \cite{bertino2008survey}. In Section \ref{sec:ExperimentalSetup}, we discuss the experimental setup that includes the specification of the data in our experiments and the architecture of our neural network. In Section \ref{sec:ExperimentalResults}, we evaluate our proposed privacy-preserving edge \gls{fl} framework in terms of overheads, complexity, latency, and privacy, based on an implementation of the proposed framework on Amazon's AWS cloud platform.

\subsection{Experimental Setup}
\label{sec:ExperimentalSetup}
\subsubsection{Epilepsy Dataset}\label{sec:database}
In this work, we consider the CHB-MIT database \cite{goldberger2000physiobank,shoeb2009application} that contains \gls{eeg} signals from 23 epilepsy patients with intractable seizures. All recordings are collected from children and young adults in the 1.5--22 age range. For collecting data, the international 10-20 system is used, which exploits \gls{eeg} electrodes with a frequency of 256 Hz and 16-bit resolution. The dataset is annotated by the medical experts and contains a total of 182 seizures. These \gls{eeg} signals are sampled at $F_s = 256$ Hz, with 16-bit resolution. Here, we consider only the two channels $T7F7$ and $T8F8$, as used in e-Glass wearable system \cite{Sopic18ISCAS}, which have been shown to be important for the detection of epileptic seizures.

\subsubsection{Seizure Detection Performance Metrics} \label{sec:Performance_Metrics}
In this work, 
we evaluate the prediction performance of our proposed privacy-preserving \gls{fl} framework based on the precision, recall, accuracy, and F1-score, defined as follows,

\noindent
\begin{align}
Precision &= \frac{TP}{ TP + FP},\\
Recall &= \frac{TP}{ TP+ FN},\\
Accuracy &= \frac{TP+TN}{ TP+FP+TN+FN},\\
F1{\text -}score &= 2 \cdot \frac{Precision \cdot Recall}{ Precision + Recall},
\end{align}
where $FP$, $TN$, $TP$, and $FN$ capture the number of false positives, true negatives, true positives, and false negatives, respectively.

\subsubsection{Deep Neural Network's Architecture} \label{sec:Network_Architecture}

As described in Section \ref{sec:Method}, we have two phases of \gls{fl}. In the first phase, since the training is performed on decentralized data stored in several hospitals, the assumption is that we do not have limitations in regard to the resources required for training our model. Therefore, we can have more trainable parameters in the network's base, if increasing the size of the base improves the classification performance. On the other hand, in the second phase, since the training is performed on resource-constrained devices, we desire to have a lower number of network parameters to train. To this end, we designed our network to have three convolution layers with 135,052 trainable parameters in the base to more effectively extract the features from input \gls{eeg} signals. For the head of our network, we merely employed one dense layer with 8,234 trainable parameters.

\begin{figure}
        \centering
            {\includegraphics[width=1.0\columnwidth]{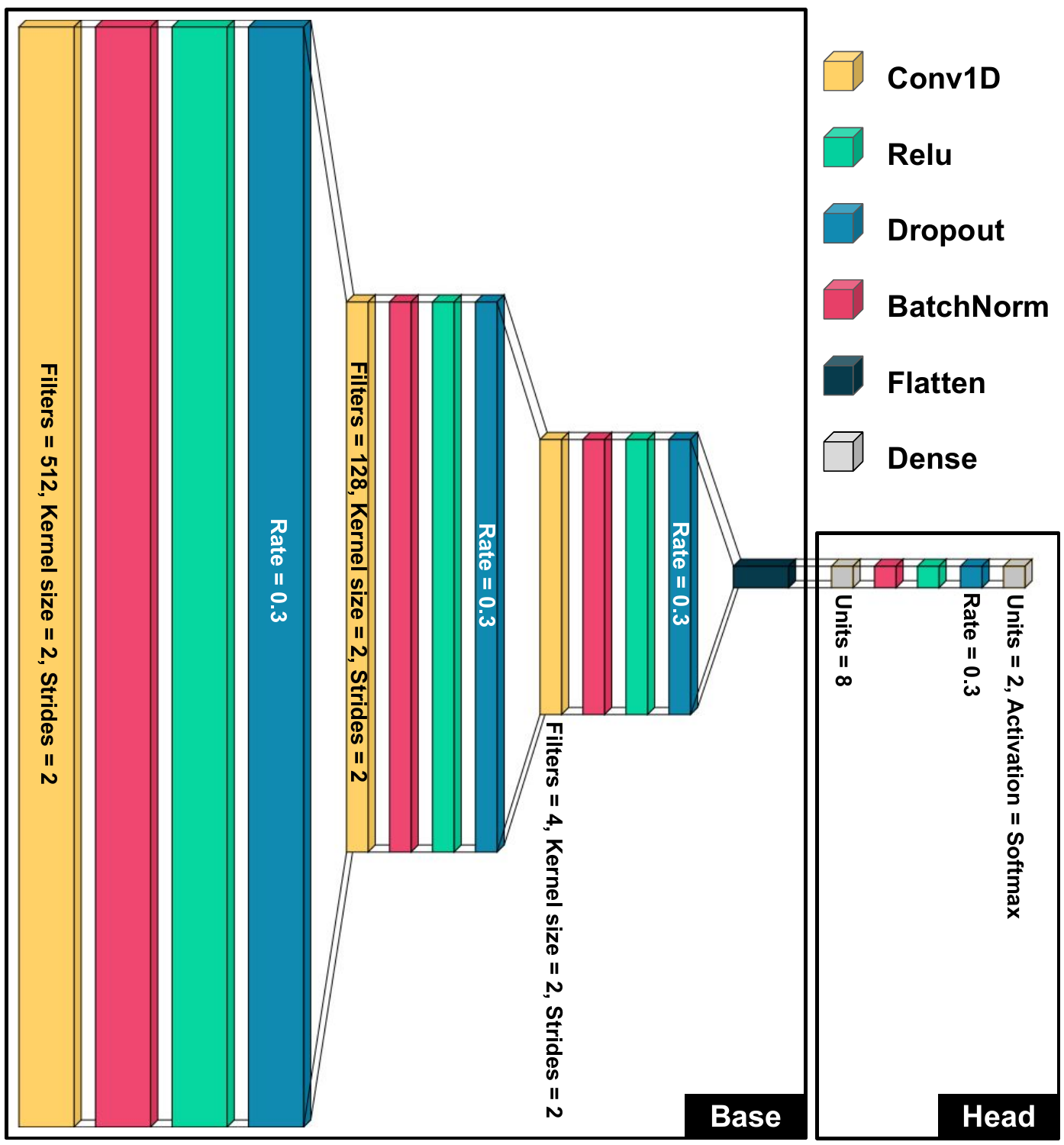}}   
        \caption[ caption ]
        {Network Architecture for Epileptic Seizure Detection\vspace{1cm}} 
        \label{fig:Network_Architecture}
\end{figure}

Figure \ref{fig:Network_Architecture} shows the architecture of the base and head of the neural network we used in our experiments.\footnote{The network architecture was plotted using \cite{Gavrikov2020VisualKeras}.}
We designed the base to have three layers of one-dimensional convolution, each followed by batch normalization, activation function, and dropout. The convolution kernel size and stride are set to 2, and the number of filters for the first, second, and third layers are set to 512, 128, and 4, respectively. The activation functions in the base are rectified linear unit (ReLu) functions, and the dropout rates are set to 30\%. For the head of the network, we consider a dense layer with eight units, followed by batch normalization, ReLu activation function, and dropout. Then, we have an output layer that contains two units with softmax activation functions. 

We implement the described network using the TensorFlow library. 
For training the artificial neural network, we have set the batch size to $16$, the learning rate to $1e^{-3}$, and the number of epochs to 50. We use the  $\rm{Adam}$ optimizer for training the network. Furthermore, we divide the dataset into three parts: training set (60\%), validation set (20\%), and test set (20\%).

\begin{table*}[h]
\caption{Quantitative comparison of the proposed framework against the baseline techniques (average of
100 runs)}
\resizebox{\textwidth}{!}{%
\setlength{\tabcolsep}{3pt}
\begin{tabular}{c c c c c c c c c}
\Xhline{3\arrayrulewidth}
\multicolumn{1}{c}{\textbf{Approach}} &
\multicolumn{1}{c}{\textbf{Electrodes}} &
\multicolumn{1}{c}{\textbf{Architecture}} &
\multicolumn{1}{c}{\textbf{Trained on Edge}} &
\multicolumn{1}{c}{\color{black}\textbf{Param. on Edge}} &
\multicolumn{1}{c}{\textbf{Precision}}&
\multicolumn{1}{c}{\textbf{Recall}}&
\multicolumn{1}{c}{\textbf{Accuracy}}&
\multicolumn{1}{c}{\textbf{F1-score}}\\
\Xhline{3\arrayrulewidth}
Baseline 1 & All & Head & Head & \color{black}8,234 (5.7\%) & 69.9\% & 65.0\% & 66.0\% & 63.3\%\\ 
Baseline 2 & T7F7+T8F8 & Head & Head & \color{black}8,234 (5.7\%) & 72.2\% & 67.0\% & 67.9\% & 65.5\%\\
Baseline 3 & All & Base+Head & Base+Head & \color{black}143,286 (100\%) & 66.7\% & 66.0\% & 65.8\% & 65.4\% \\ 
Baseline 4 & T7F7+T8F8 & Base+Head & Base+Head & \color{black}143,286 (100\%) & 83.8\% & 82.7\%& 82.9\%& 82.7\% \\ 
Our Framework & T7F7+T8F8 & Base+Head & Head & \color{black}8,234 (5.7\%) & 82.1\% & 81.2\%  & 81.4\% & 81.2\%\\ 
\color{black} \begin{tabular}[c]{@{}c@{}}Our Framework\\ (personalized) \end{tabular} & \color{black} T7F7+T8F8 & \color{black} Base+Head & \color{black} Personalized Head & \color{black}8,234 (5.7\%) &  \color{black}88.7\% & \color{black}86.1\%  & \color{black}88.4\% & \color{black}87.4\%  \\
\Xhline{3\arrayrulewidth}
\end{tabular}
}
\label{tab:results_performance}
\end{table*}

\subsection{Experimental Results} \label{sec:eval_quality}
\label{sec:ExperimentalResults}

\subsubsection{Prediction Performance}

In this section, we evaluate our proposed edge \gls{fl} framework in terms of prediction performance and for the case of epileptic seizure detection. We perform the experiments 100 times and report the average value in Table \ref{tab:results_performance}. \color{black}Consistent with the CHB-MIT dataset, we consider 23 patients and each with his/her own data locally and the e-Glass system \cite{Sopic18ISCAS}. \color{black} We consider four baseline approaches as follows:
\begin{itemize}
    \item \emph{Baseline 1:} In this baseline, we only consider the head architecture trained on data from all \gls{eeg} electrodes, but evaluated based only on the electrodes present in the e-Glass wearable system, i.e., T7F7 and T8F8. Our proposed framework outperforms Baseline 1 by a large margin, i.e., more than $15\%$, both in terms of accuracy and F1-score. These results show the importance of a pre-trained base architecture and re-training the head architecture with the data from the e-Glass wearable system.
    \item \emph{Baseline 2:} In this baseline, we only consider the head architecture trained and evaluated based on the electrodes present in the e-Glass wearable systems, i.e., T7F7 and T8F8. Our proposed framework outperforms Baseline 2 by a large margin, i.e., $13.5\%$ in terms of accuracy and $15.7\%$ in terms of F1-score. These results show the impact of exploiting a pre-trained base architecture, even if it is not based on the data from the e-Glass wearable system.
    \item \emph{Baseline 3:} In this baseline, we consider the entire neural network architecture, both the base architecture and the head architecture, trained on data from all \gls{eeg} electrodes, but evaluated based only on the electrodes present in the e-Glass wearable systems, i.e., T7F7 and T8F8. Our proposed framework outperforms Baseline 3 by a large margin, i.e., more than $15\%$, both in terms of accuracy and F1-score. These results show the impact of re-training the head architecture with the data from the e-Glass wearable system.
    \item \emph{Baseline 4:} In this baseline, we consider the entire neural network architecture, both the base architecture and the head architecture, trained and evaluated based on the electrodes present in the e-Glass wearable systems, i.e., T7F7 and T8F8. The small gap between Baseline 4 and our proposed framework, i.e., less than $1.5\%$, demonstrates the relevance of our proposed framework in the context of resource-constrained wearable and mobile-heath technologies, dramatically reducing the training overheads (because our framework only requires training the head architecture, for which the number of parameters to be trained is less than $6\%$ of the entire architecture, as it will be shown in the next section.), with only a marginal loss in terms of prediction performance. 
\end{itemize}

The experiments in this section show the superiority of our proposed framework against Baseline 1, Baseline 2, and Baseline 3, in terms of performance prediction.  At the same time, we show that our proposed framework reaches almost the same quality (less than $1.5\%$ difference) as training the entire network based on the data from the e-Glass wearable system. However, as we will show in the next section, training the entire network is practically infeasible on resource-constrained mobile-health systems.

\color{black}
Next, our proposed framework also provides the possibility of personalization. In this setting, we follow the same exact procedure as discussed in our proposed framework. After the \gls{fl} is finalized, we fine-tune the head architecture with the local data of the patient to develop a personalized model for that patient. We have repeated this experiment for all patients to obtain a personalized model for each patient.  The last row of Table \ref{tab:results_performance} shows the average results of the personalization for all patients. The results indicate that the personalized models outperform their generic counterpart by $7.0\%$ and $6.2\%$ in terms of accuracy and F1-score, respectively. This demonstrates that our proposed framework supports personalization to significantly improve the prediction performance. 
\color{black}

\begin{table}[]
\caption{\color{black}
Evaluation of the impact of network architectures (average of 10 runs)\color{black}}
\resizebox{\columnwidth}{!}{\color{black}
\setlength{\tabcolsep}{2pt}
\begin{tabular}{lcccccc}
\Xhline{3\arrayrulewidth}
\textbf{Architecture}                         & \textbf{Approach}  & \textbf{Param. on Edge}    & \textbf{Precision} & \textbf{Recall} & \textbf{Accuracy} & \textbf{F1-score} \\ \Xhline{3\arrayrulewidth}
\multirow{2}{*}{Transformer \cite{vaswani2017attention}}  & Baseline 4    & 128,882 &      84.1\%        &     81.9\%       &       83.4\%       &       82.3\%       \\
                             &                    Our Framework & 8,234 &       80.3\%        &     79.8\%       &       80.0\%       &       79.9\%       \\ \hline
\multirow{2}{*}{ResNet \cite{he2016deep}}    & Baseline 4    &  145,226 &     84.8\%        &     82.6\%       &       80.9\%       &       81.0\%       \\
                             &                    Our Framework & 8,234 &       83.9\%        &     80.3\%       &       80.3\%       &       80.6\%      \\ \Xhline{3\arrayrulewidth}
\end{tabular}}
\label{tab:DNNarchitectures}
\end{table}

\color{black}
Finally, here, we show that the proposed approach is not limited to the architecture considered in Figure \ref{fig:Network_Architecture}. To demonstrate this, we consider two state-of-the-art architectures, namely Transformer \cite{vaswani2017attention} and ResNet \cite{he2016deep}. Our Transformer model consists of one encoder block, including multi-head attention with 8 heads and an embedding size of 32. The feedforward part consists of two 1-dimensional convolutional layers with kernel sizes of 1, with 8 and 16 filters, and strides of 2. Each layer consists of 1-dimensional max pooling with a size of 2, followed by a batch normalization layer and ReLU activation functions. Global average pooling is performed on the output of the encoder. Our ResNet model consists of one convolutional residual block, including two 1-dimensional convolutional layers with kernel sizes 4 and 2, with 128 and 4 filters, with strides of 4 and 2, respectively. The main block is comprised of three 1-dimensional convolutional layers, each with a kernel size of 3 with 512, 128, and 4 filters. Each layer is followed by a batch normalization layer and ReLU activation functions.

In Table \ref{tab:DNNarchitectures}, we compare our scheme in terms of the performance against Baseline 4 (same settings as Baseline 4 in Table 1), where the entire architecture is trained on the edge. Let us first consider the Transformer model. In Baseline 4, both base and head architectures are trained on the edge, i.e., a total of 128,882 parameters. This results in an accuracy and an F1-score of $83.4\%$ and $82.3\%$, respectively. When using our approach, however, we reduce the number of parameters trained on the edge to 8,234, which is roughly $6.4\%$ of the entire architecture, while obtaining a similar performance. In the case of ResNet model, Baseline 4 trains the entire architecture on the edge, which has a total of 145,226 parameters. This results in an accuracy and an F1-score of $80.9\%$ and $81.0\%$, respectively. When using our approach, however, we reduce the number of parameters trained on the edge to 8,234, which is roughly $5.7\%$ of the entire architecture, while obtaining on par performance (less than $1\%$ difference). This evaluation across diverse architectures underscores the adaptability and effectiveness of our proposed approach, showcasing its applicability beyond a specific network configuration considered earlier.
\color{black}

\begin{table}[t]
\caption{\color{black}
Evaluation of the impact of knowledge distillation with several student network architectures (average of 10 runs)\color{black}}
\resizebox{\columnwidth}{!}{\color{black}
\setlength{\tabcolsep}{2pt}
\begin{tabular}{lccccc}
\Xhline{3\arrayrulewidth}
 \textbf{Model} & \textbf{Parameters} &\textbf{Precision} & \textbf{Recall} &\textbf{Accuracy} & \textbf{F1-score}
\\
\Xhline{3\arrayrulewidth}

Teacher & {540,726} &83.8\% & 83.5\% & 83.7\% & 83.5\%\\

Student 1 & {275,766} &82.9\% & 82.6\% & 82.6\% & 82.6\% \\

Student 2 & {143,286} &82.1\% & 81.2\% & 81.4\% & 81.2\% \\

Student 3 & {76,470} &79.7\% & 79.5\% & 79.5\% & 79.4\% \\ \Xhline{3\arrayrulewidth} 
\end{tabular}}
\label{tab:distil}
\end{table}

\begin{figure*}[]
        \centering
        \begin{minipage}{\textwidth}
        
        \begin{subfigure}[b]{.33\textwidth}
        \includegraphics[width=\textwidth]{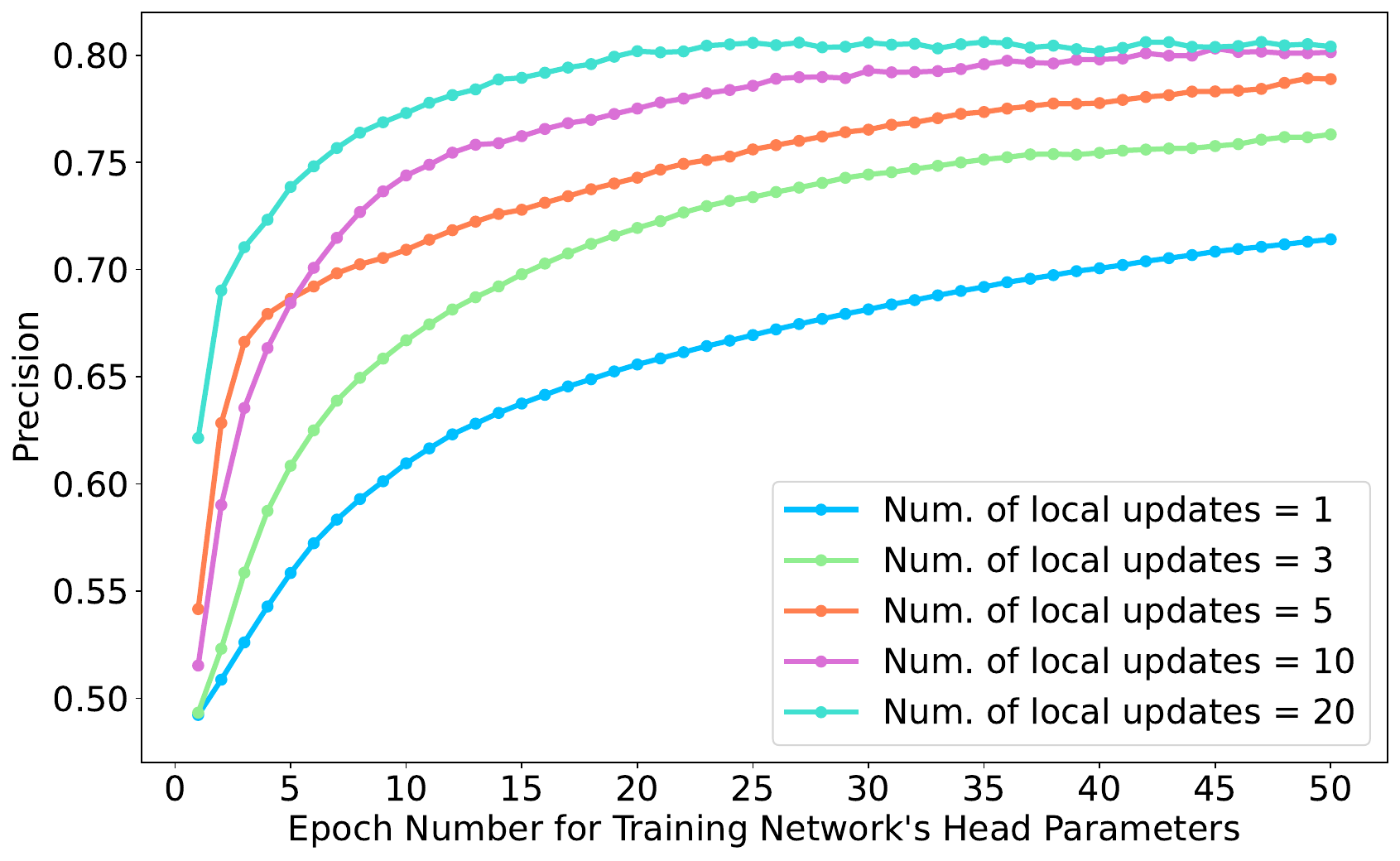}
        
        \label{fig:Exp110}
        \end{subfigure}
        \begin{subfigure}[b]{.33\textwidth}
        \includegraphics[width=\textwidth]{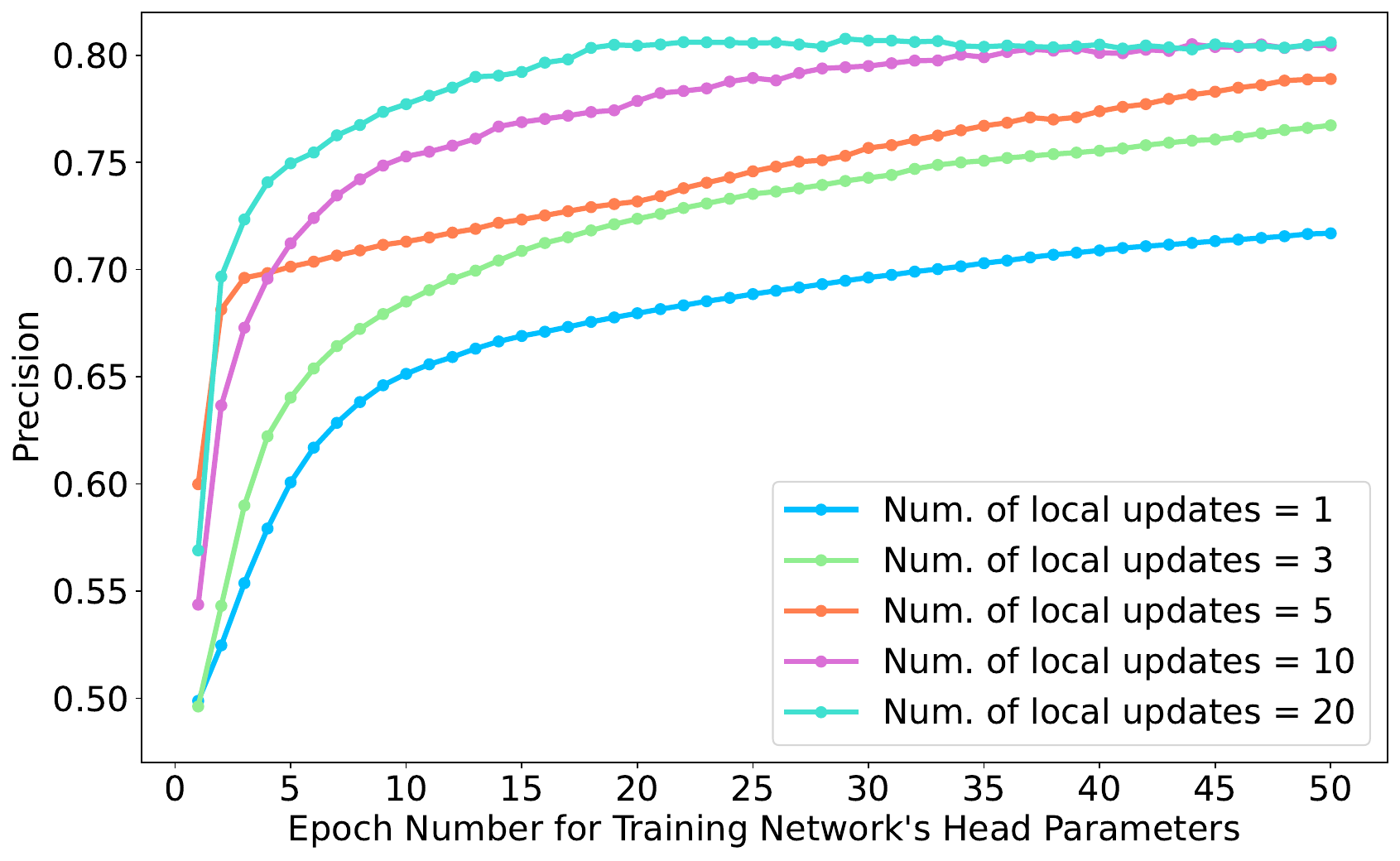}
  
        \label{fig:Exp120}
        \end{subfigure}
        \begin{subfigure}[b]{.33\textwidth}
        \includegraphics[width=\textwidth]{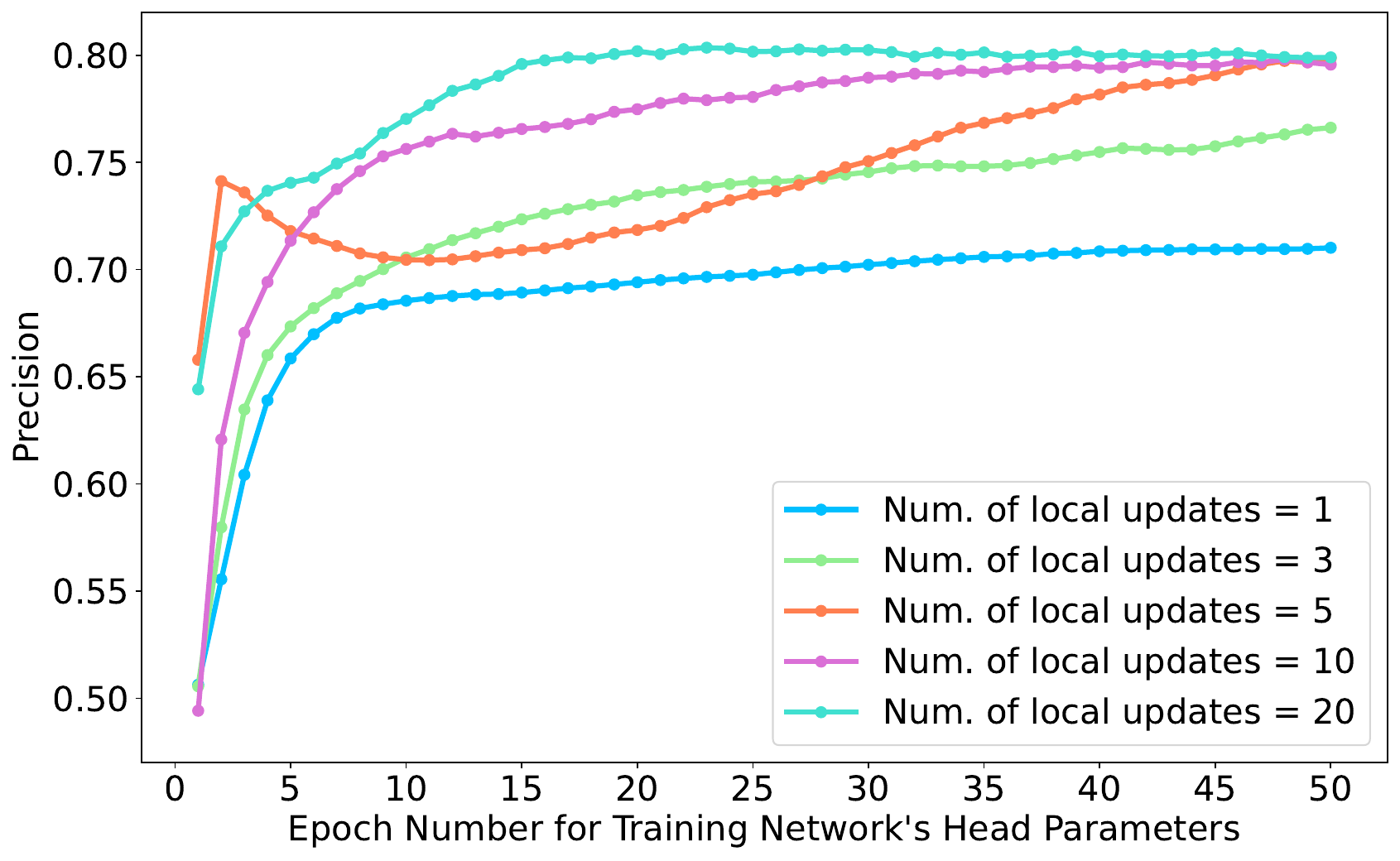}

        \label{fig:Exp20}
        \end{subfigure}
        
        \vskip\baselineskip
        
        \begin{subfigure}[b]{.33\textwidth}
        \includegraphics[width=\textwidth]{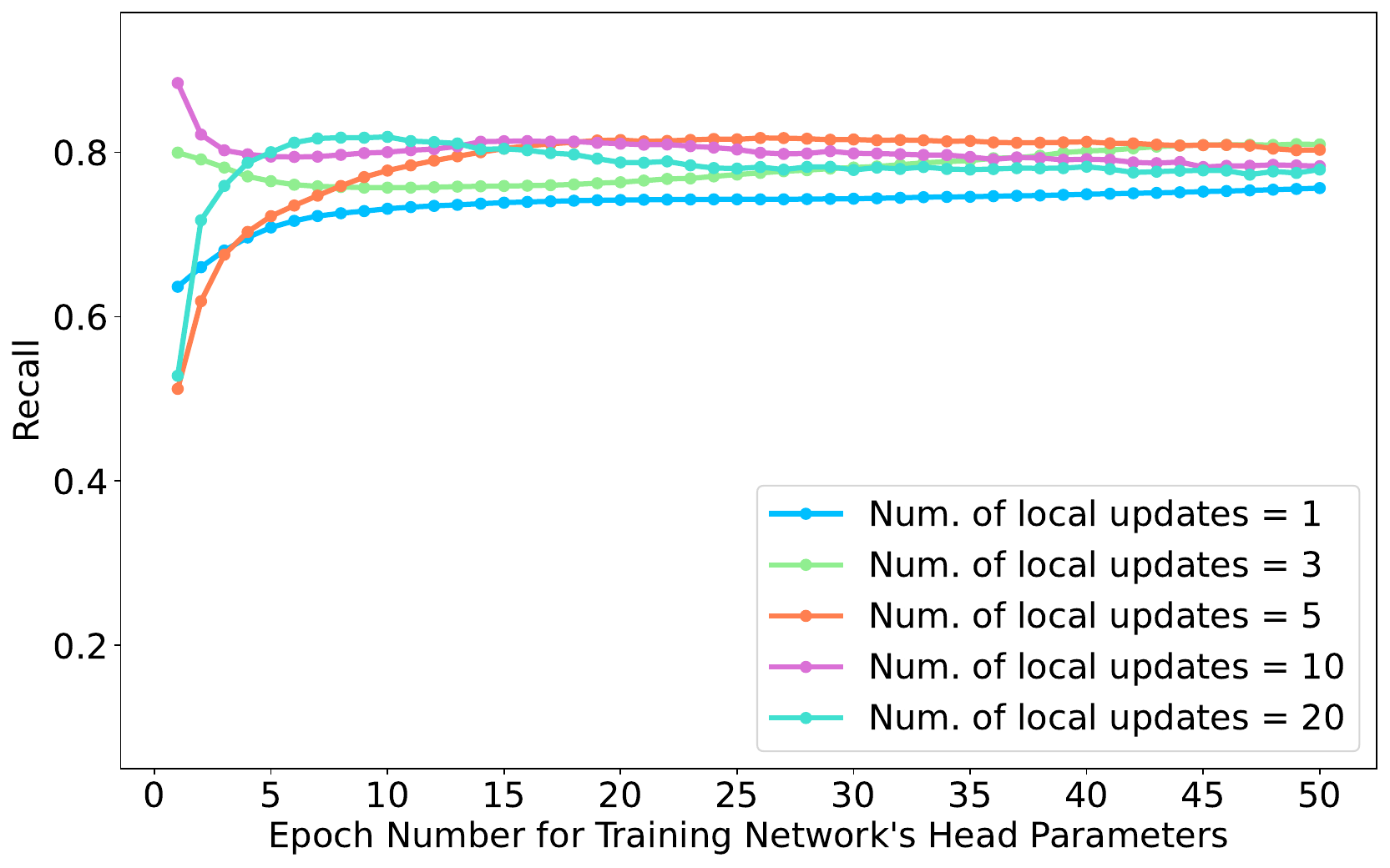}
        
        \label{fig:Exp111}
        \end{subfigure}
        \begin{subfigure}[b]{.33\textwidth}
        \includegraphics[width=\textwidth]{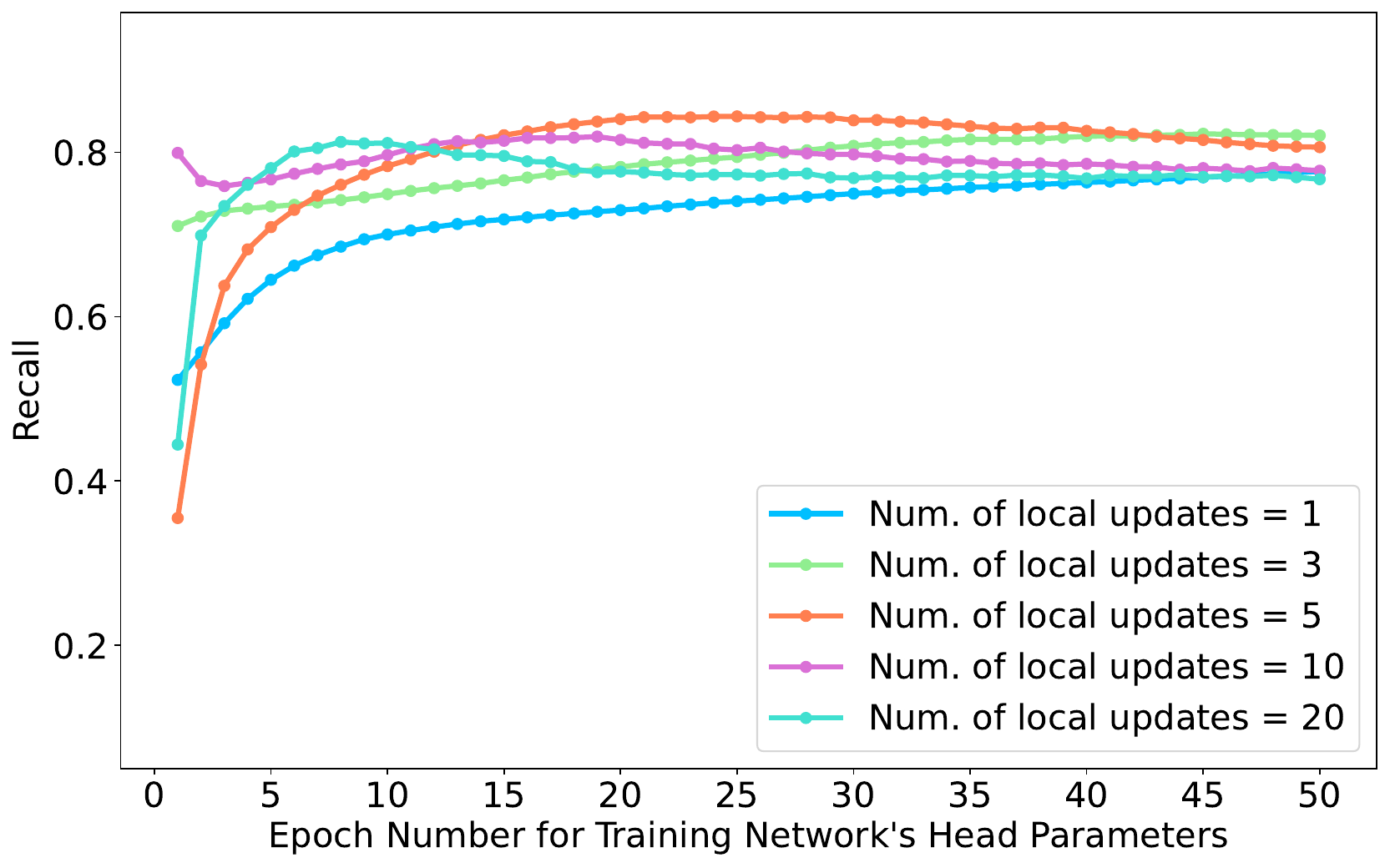}
  
        \label{fig:Exp121}
        \end{subfigure}
        \begin{subfigure}[b]{.33\textwidth}
        \includegraphics[width=\textwidth]{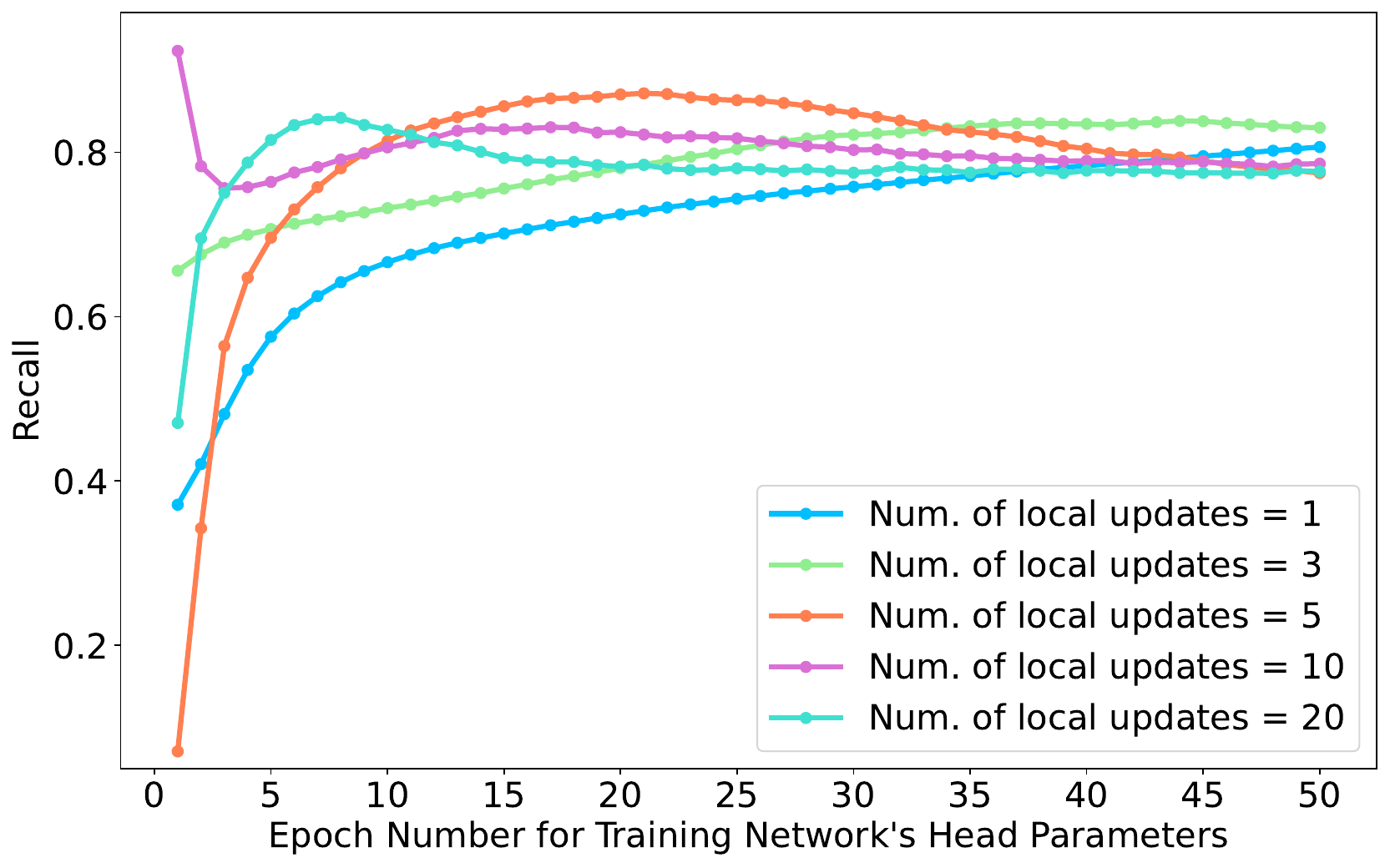}

        \label{fig:Exp21}
        \end{subfigure}
        
        \vskip\baselineskip

        \begin{subfigure}[b]{.33\textwidth}
        \includegraphics[width=\textwidth]{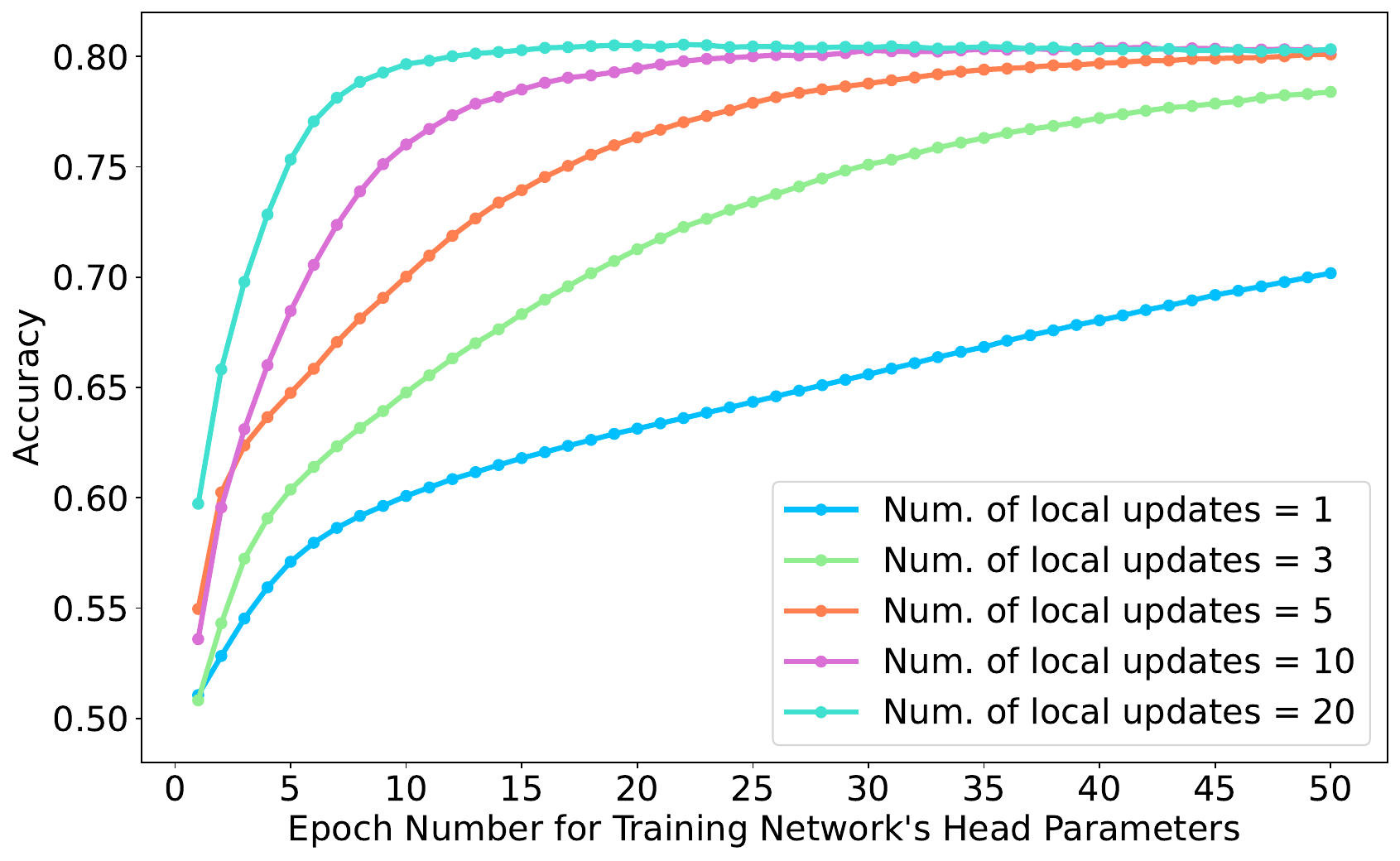}
        
        \label{fig:Exp112}
        \end{subfigure}
        \begin{subfigure}[b]{.33\textwidth}
        \includegraphics[width=\textwidth]{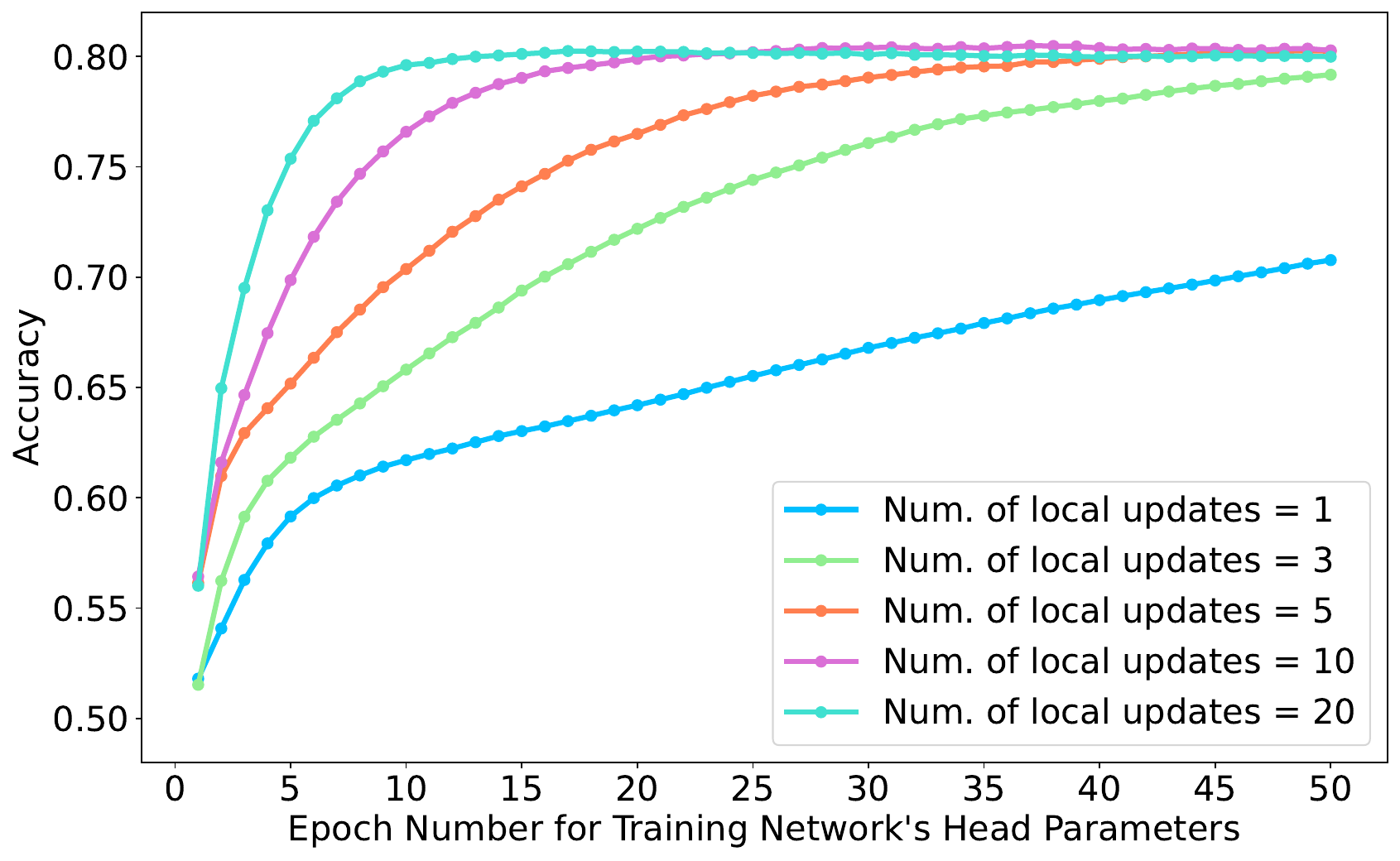}
  
        \label{fig:Exp122}
        \end{subfigure}
        \begin{subfigure}[b]{.33\textwidth}
        \includegraphics[width=\textwidth]{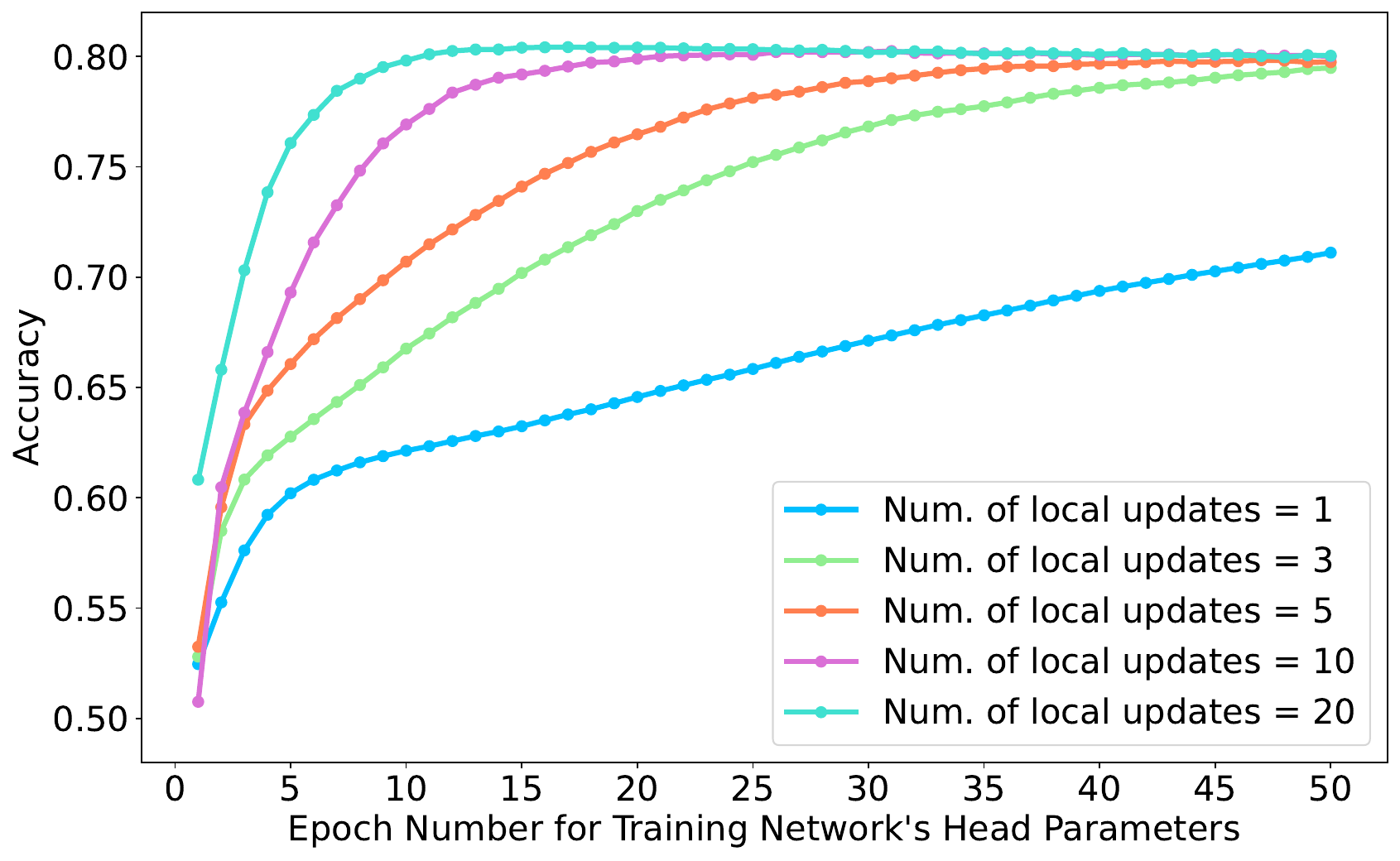}

        \label{fig:Exp22}
        \end{subfigure}
        
        \vskip\baselineskip
        
        \begin{subfigure}[b]{.33\textwidth}
        \includegraphics[width=\textwidth]{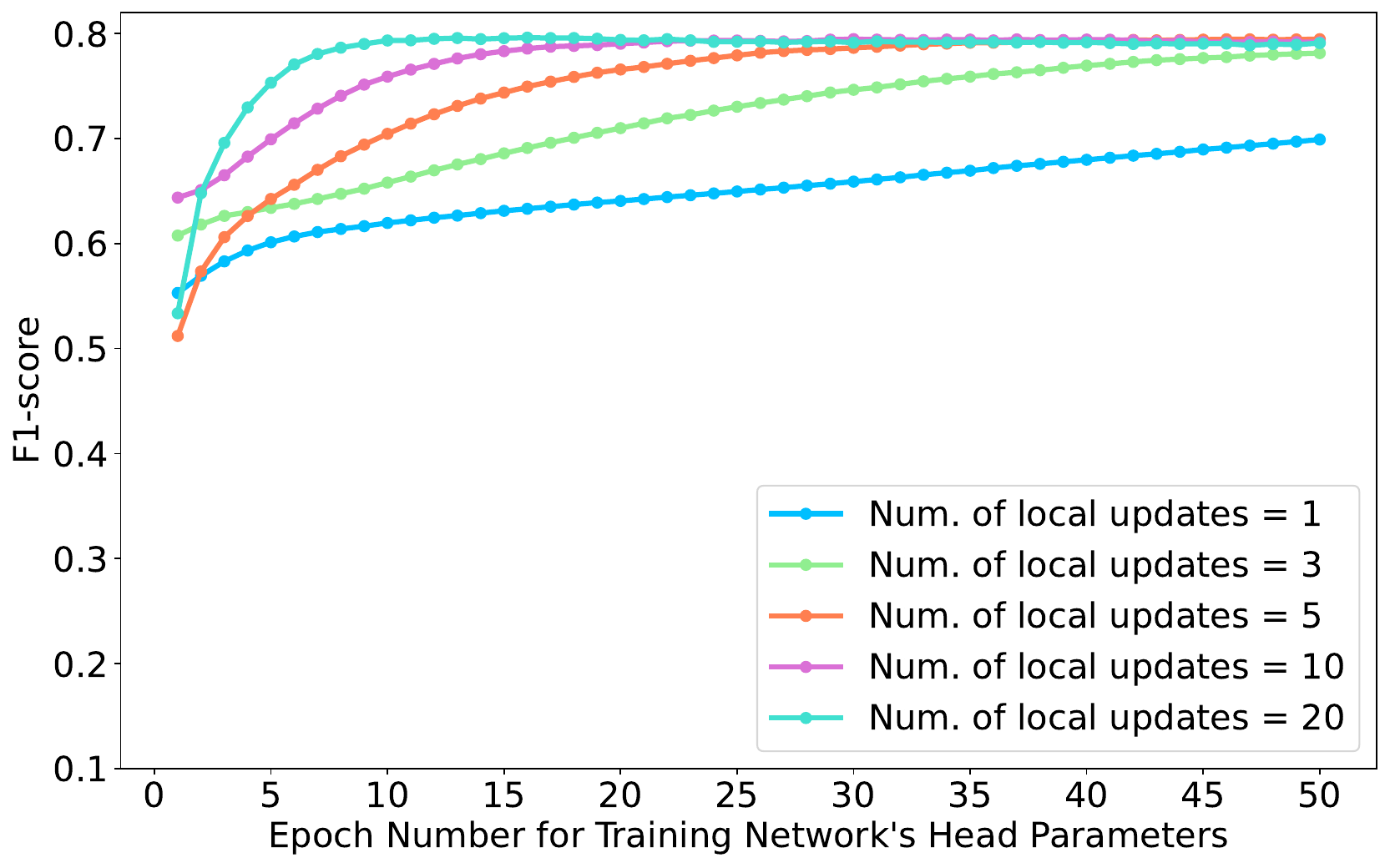}
        \caption[part 1]
        {{\small Scenario one}}
        \label{fig:Exp114}
        \end{subfigure}
        \begin{subfigure}[b]{.33\textwidth}
        \includegraphics[width=\textwidth]{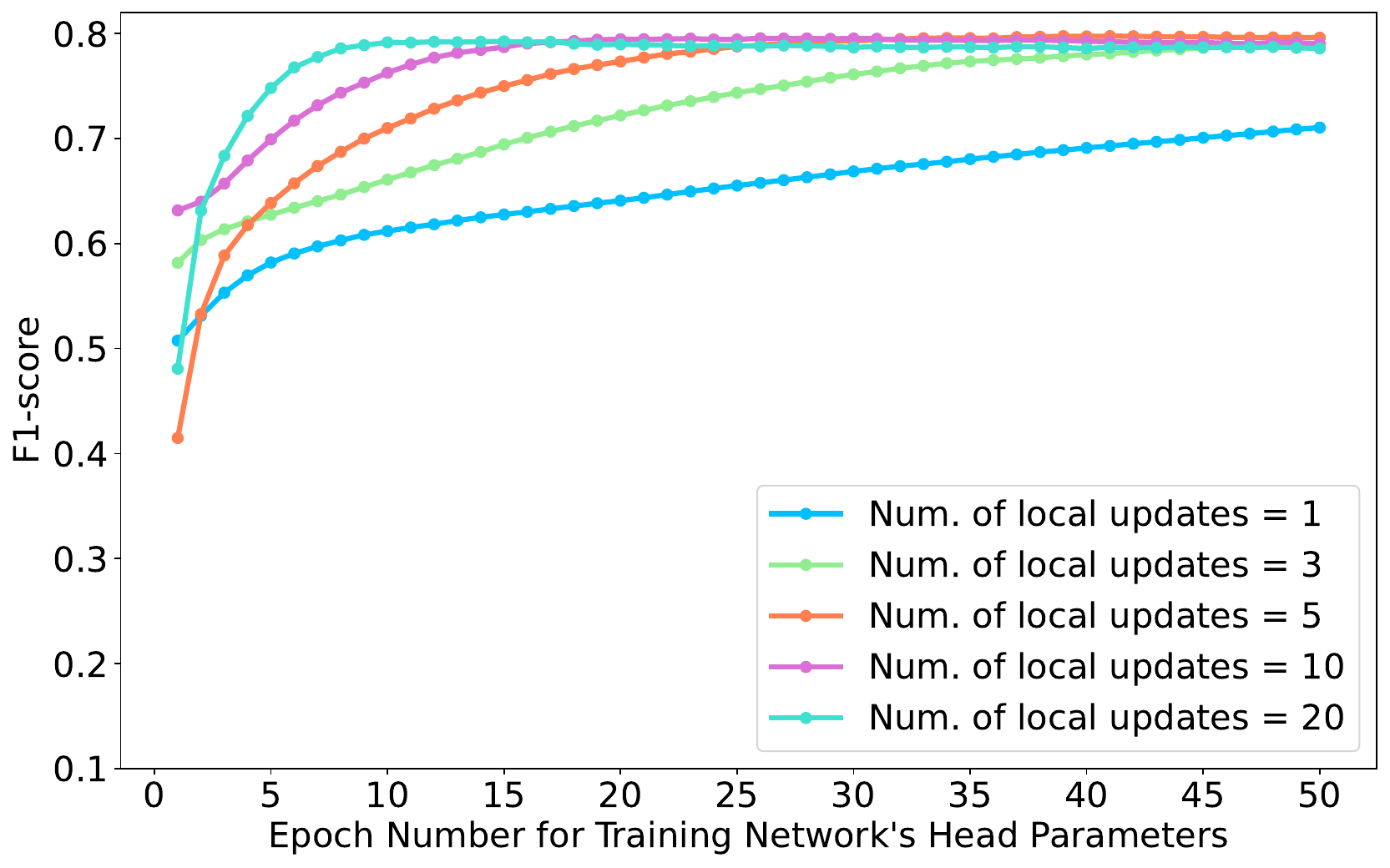}
        \caption[part 2]
        {{\small Scenario two}}
        \label{fig:Exp123}
        \end{subfigure}
        \begin{subfigure}[b]{.33\textwidth}
        \includegraphics[width=\textwidth]{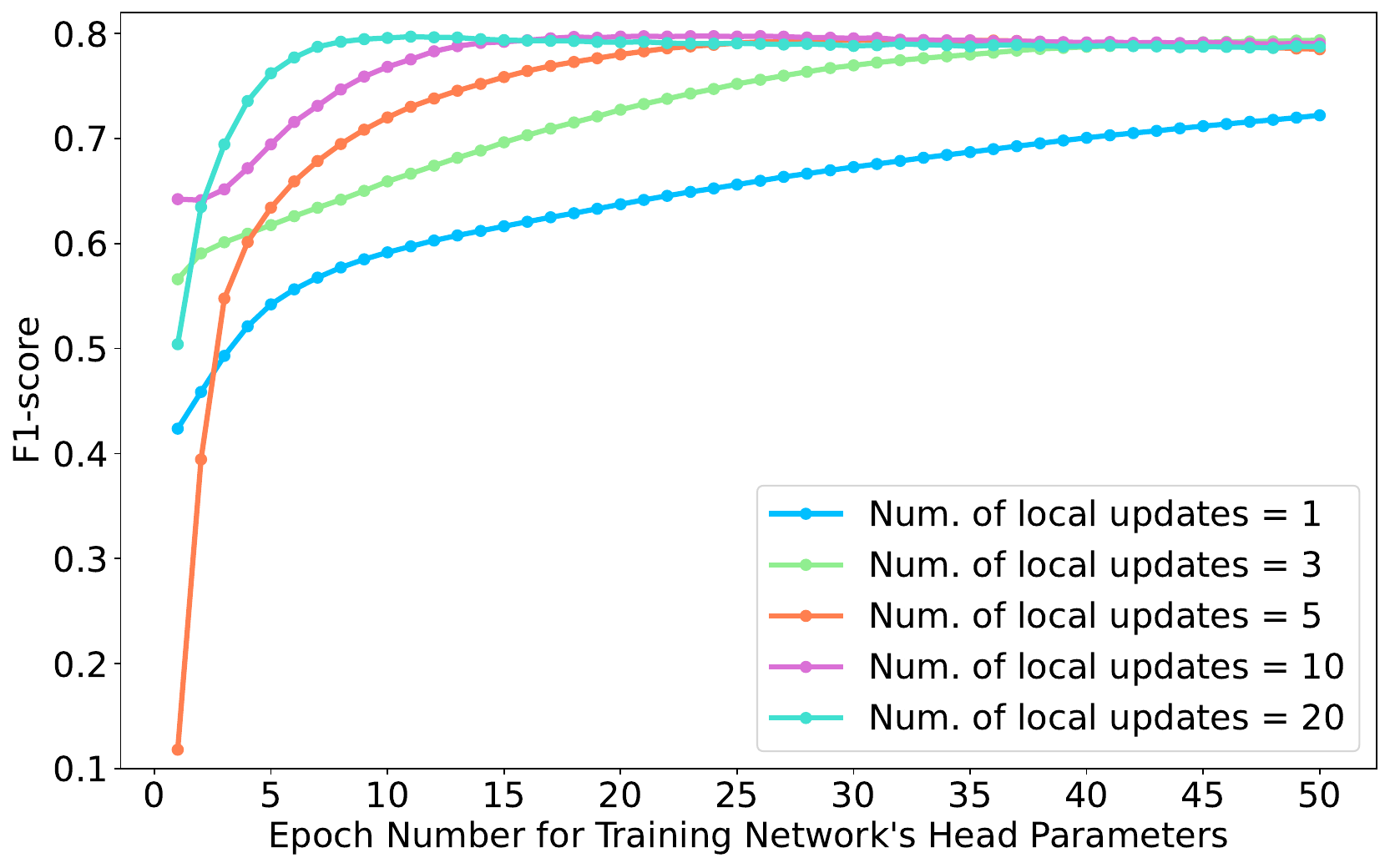}
        \caption[Part 3]%
        {{\small Scenario three}}  
        \label{fig:Exp23}
        \end{subfigure}
        
        \end{minipage}

        \caption[ caption ]
        {\small The impact of locally updating the network's head's parameters on classification performance for Scenarios one, two, and three (average of 100 iterations)}
         \label{fig:local_updates}
\end{figure*}
\subsubsection{Performance and Overhead}
\label{sec:Performance_Overhead}

In this section, we evaluate the overhead of our proposed edge \gls{fl} framework.

\color{black}
Let us first look into the impact of knowledge distillation for several student network architectures. The results are shown in Table \ref{tab:distil}. Each number represents the average of 10 runs. The results show the trade-off between the size of the student network architecture, which is captured by the number of parameters, and the prediction performance metrics, captured by the precision, recall, accuracy, and F1-score. We have selected \emph{Student 2} as a favorable trade-off between the network size and prediction performance and reported the results in Table \ref{tab:results_performance}, which will be discussed further in this section.\color{black}

As described in Section \ref{sec:Network_Architecture}, the base and head architectures in the \emph{Student 2} network have three layers (with 135,052 trainable parameters) and one layer (with 8,234 trainable parameters), respectively. Our results show that, it is possible to reach a prediction performance of $81.2\%$ by training less than $6\%$ of the parameters in the entire deep neural network on the edge IoT device, compared to training the entire deep neural network with $82.7\%$. Therefore, our proposed framework enables efficient training of neural networks on resource-constrained IoT devices, without any major loss in terms of prediction performance. 

Next, we look into how increasing the number of times each data-holder party locally updates the network's head parameters can reduce the communication overheads in the training process. 
We perform our experiments based on three scenarios. In Scenarios one, two, and three, we, respectively, have 3, 5, and 10 data-holder parties. The training data 
is evenly distributed among data-holder parties in each scenario. We increase the number of local updates from 1 to 20 to evaluate how many epochs it takes in each scenario to obtain a certain classification performance. The higher the number of epochs it takes to get the desired performance result, the higher the communication overheads imposed in the training process, and the higher the energy overheads of \gls{fl}.

Figure \ref{fig:local_updates} shows the results of the experiments. As the results show, if we locally update the head parameters only once at each epoch, we converge to the global optimum slowly. Slow convergence means that we require more epochs and rounds of secure aggregation to train our model. This increases the communication overheads and is not desirable in the context of resource-constraint battery-powered devices. On the other hand, we observe that by increasing the number of local updates to 3, 5, 10, and 20, the convergence rate is faster. Among these, the results for 20 local updates are the best. For Scenarios one, two, and three, it takes approximately 10 epochs to converge when we consider 20 local updates. As shown in Figure \ref{fig:local_updates}, this means at least 40 fewer epochs for convergence in comparison to 1 local update, because the convergence is not achieved even after 50 epochs when we consider 1 local update.

\subsubsection{Privacy and Overhead}
\label{sec:Privacy_Overhead}

In this section, we discuss the communication overhead, communication latency, and privacy of our \gls{smc} technique. We compare our techniques against three baseline \gls{smc} approaches, i.e., NOSMC, STSMC, and Shamir \cite{10.1145/359168.359176}. Table \ref{tab:SMC_eval} presents the analysis results for all the techniques. In this table, the latency/delay of transferring a packet from $P_i$ to $P_j$ is denoted by $L(P_i, P_j)$.

\begin{table*}[]
\caption{Communication complexity, communication latency and privacy of \gls{smc} approaches ($n=$ number of data holders, $N=$ number of all parties including the mediator)}
\resizebox{\textwidth}{!}{\begin{tabular}{@{}ccccccc@{}}
\Xhline{3\arrayrulewidth}
\multirow{2}{*}{\textbf{Approach}}                & \multirow{2}{*}{\textbf{Party}}  & \multicolumn{2}{l}{\textbf{Communication}}              & \textbf{Communication} & \textbf{Communication Latency} & \textbf{Min. Number of} \\ \cline{3-4}
 &                                                   & \textbf{Send}               & \textbf{Receive}            &       \textbf{Complexity}                  &             \textbf{(M = Mediator)}       & \textbf{Colluding Parties}        \\ \Xhline{3\arrayrulewidth}
\multirow{2}{*}{NOSMC} & Data Holders & $1$ & $0$ & \multirow{2}{*}{$O(n)$}                & \multirow{2}{*}{$\max_{i} L(P_i, M),\ i \in \{1, \dots, n\}$} & \multirow{2}{*}{1}      \\
                        & Mediator      & $0$ & $N-1$                                      &                                         &                             \\ \midrule
STSMC       & All         & $2$ & $2$                             & $O(n)$                                     & {$2 \cdot (\sum_{i=1}^{n-1} L(P_i, P_{i+1})+L(P_n, P_1)) $}  & 2  \\ \midrule

\multirow{3}{*}{Shamir \cite{10.1145/359168.359176}}                & $k-1$ Parties         & \multicolumn{1}{c}{$n$} & \multicolumn{1}{c}{$n-1$} & \multirow{3}{*}{$O(n^2)$}       & \multirow{1}{*}{$\max_{i,j} L(P_i, P_j), i,j \in \{i,j \in \{1, \dots, n\} \mid i \neq j\}$}              & \multirow{3}{*}{$k$}    \\
 & $1$ Party         & \multicolumn{1}{c}{$n-1$} & \multicolumn{1}{c}{$n+k-2$} & & \multirow{1}{*}{$+$} & \\
  & The Rest         & \multicolumn{1}{c}{$n-1$} & \multicolumn{1}{c}{$n-1$} & & \multirow{1}{*}{$\max_{i} L(P_i, M), i \in \{1, \dots, n\}$}&  \\ \midrule
  
\multirow{2}{*}{Proposed Framework} & Data Holders & 1 & 0 & \multirow{2}{*}{$O(n)$}                & \multirow{2}{*}{$\max_{i} L(P_i, M), i \in \{1, \dots, n\}$}   & \multirow{2}{*}{$k$}    \\
                        & Mediator      & $0$ & $N-1$                                      &                                         &                             \\ \Xhline{3\arrayrulewidth}
\end{tabular}}
\label{tab:SMC_eval}
\end{table*}

\begin{itemize}
    \item \emph{NOSMC:} In NOSMC, we do not adopt an \gls{smc} scheme and directly share our secret values, here the array of network gradients or weights, with the mediator. This approach is a baseline for comparing different \gls{smc} schemes with respect to communication complexity, communication latency, and privacy.
    
    In regard to communication complexity, each data-holder party sends one message to the mediator. This is the minimum number of messages required for the aggregation of secret values among the presented approaches. Therefore, the communication overhead is of $O(n)$. 
    In regard to communication latency, since all data holders directly send their messages to the mediator in parallel, the latency is equal to the delay of the last message received by the mediator. 
    In regard to privacy, since secret values are directly shared with the mediator, the mediator knows about the secrets without collusion. Therefore, the minimum number of colluding parties is equal to one.

    \item \emph{STMSC:} In straightforward \gls{smc} or STSMC, we have two rounds of message passing among parties in a ring network topology. In the first round, each party adds a mask to its secret value and aggregates it with the result received from the previous party. Then, it passes the result of aggregation to the next party. In the second round, each party subtracts its mask from the aggregated result of the previous round.
    
    In regard to communication complexity, each party sends and receives one message in each round. In total, parties send/receive messages $4n$ times to obtain the aggregation result of secret values, hence the communication overhead is of $O(n)$. 
    In regard to communication latency, in this technique, at each round of secure aggregation, we have two rounds of message passing among the data-holder parties. And, all communications are performed sequentially. 
    In regard to privacy, for each party, if its two neighboring (before and after in the ring network topology) parties collude, they can reveal the party's secret value.

    \item \emph{Shamir \cite{10.1145/359168.359176}:} In Shamir's scheme, we have two rounds of message passing. In the first round, each party calculates initial results based on its secret value and coefficients and the public values of other parties. Then, each party sends and receives one message, each containing one initial result, to and from all other parties. Next, $k$ parties calculate intermediate results based on the received initial results from other parties and their own private information. In the second round, if $k$ parties share their intermediate results, then, we can determine the value of aggregation of secret values by solving a system of equations.
    
    In regard to communication complexity, in the first round, all parties holding secret values ($n$ parties) send and receive $n-1$ messages. In the second round, $k-1$ parties send one message to $1$ party, and accordingly, that party receives $k-1$ messages. The total number of receiving and sending messages equals $2(n^2-n+k-1)$. Therefore, the communication overhead is of $O(n^2)$. 
    In regard to communication latency, in the first round, all parties holding secret values send other parties one message in parallel. In the second round, in the case where $k=n$, all parties holding secret values send one message to the mediator in parallel. The first and second rounds are performed sequentially. Therefore, the latency is equal to the delay of the last message received in the first round plus the delay of the last message received by the mediator in the second round. 
    In regard to privacy, each party has $k$ private values, i.e., $k-1$ secret coefficients and $1$ secret value. If $k$ parties that received the initial results from one victim party collude in the first round, then, they can solve a system of $k$ equations ($k$ equations, $k$ unknowns) to reveal the victim party's secret value.

    \item \emph{Proposed Framework:} In our proposed framework, every data-holder party generates and aggregates masks based on the shared random seeds. Then, it shares its masked secret value with the mediator.
    
    In regard to communication complexity, similar to the NOSMC approach, each data-holder party merely sends one message to the mediator, and the communication overhead is of $O(n)$. 
    In regard to communication latency, similar to NOSMC, all data holders directly send their messages to the mediator in parallel. Therefore, the latency is equal to the delay of the last message received by the mediator. 
    In regard to privacy, in the worst case, $k$ ($k<n$) data holders who shared seeds for generating random masks with the victim party should collude to reveal a secret value. Moreover, since only the mediator receives the masked secret value, its participation in the collusion is necessary for revealing the secret value.

\end{itemize}
    
As the results in Table \ref{tab:SMC_eval} show, our proposed framework's communication complexity is $O(n)$ and similar to the NOSMC approach. This is particularly desirable for use when we are training a model on resource-constrained devices. If we had adopted Shamir's scheme, at each round of secure aggregation, each edge device, at least, should have sent and received $n-1$ messages. This is while by adopting our secret sharing scheme, each edge device, at each round of secure aggregation, should merely send one message, which is considerably more efficient than Shamir's technique. Our proposed framework's communication latency is also similar to the NOSMC approach. This is because both approaches have one round of message passing that is performed in parallel. Moreover, the privacy of our framework is similar to Shamir's scheme, i.e., in both approaches, it takes at least $k$ data-holder parties to reveal a secret value.

In summary, the analysis results show that, in terms of communication complexity and communication latency, our proposed framework is as efficient as the NOSMC approach. At the same time, our framework protects the privacy of data holders similar to Shamir's scheme.

\subsubsection{Implementation on Amazon AWS Cloud}
\begin{figure}
        \centering
            {\includegraphics[width=0.8\columnwidth]{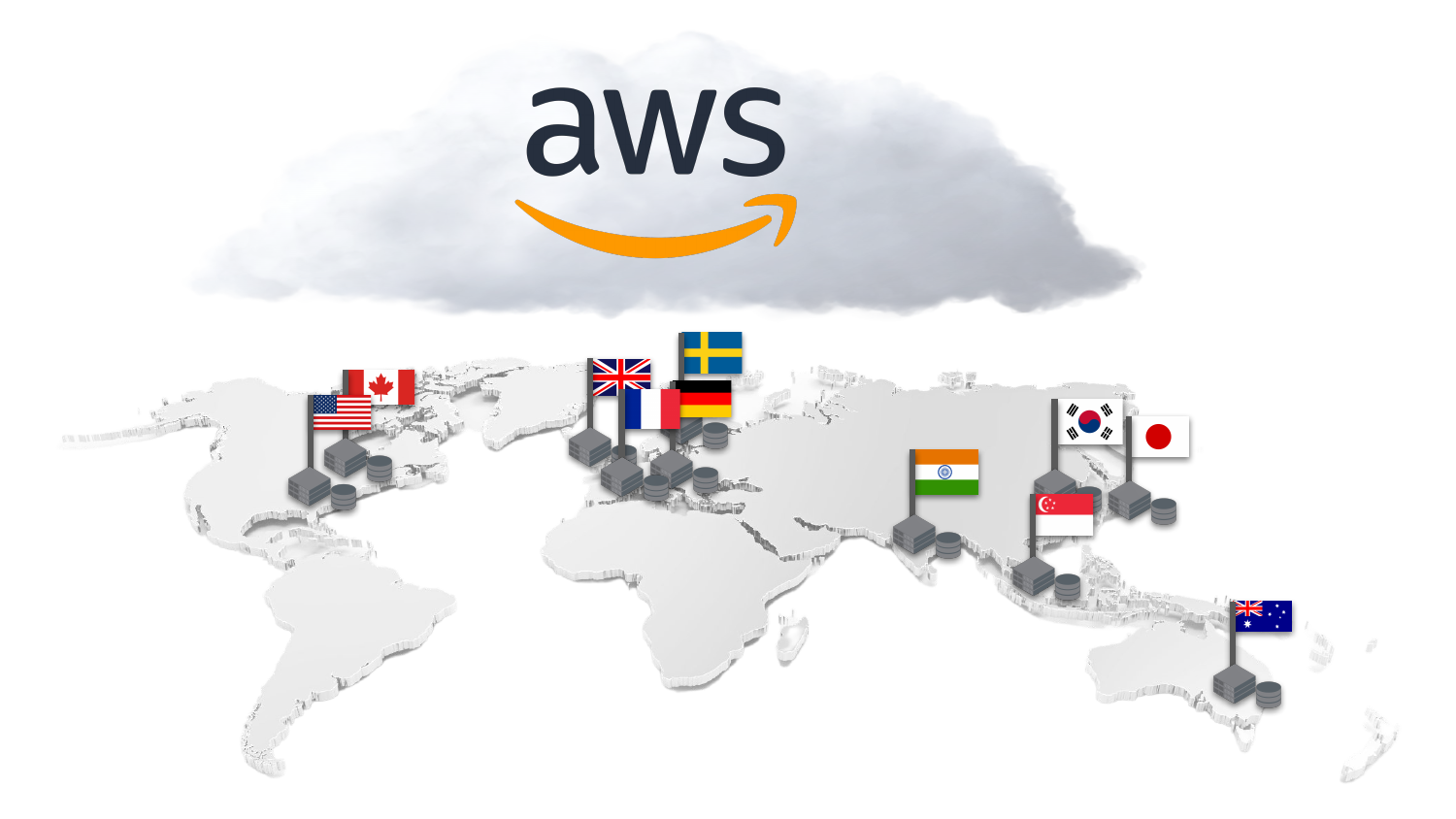}}   
        \caption[ caption ]
        {Locations of 11 Amazon’s AWS machines that serve as the resources for the mediator and data-holder parties in our experiments} 
        \label{fig:Locations_AWS}
\end{figure}

As a final evaluation step, we implement our proposed framework on Amazon's AWS cloud platform \cite{AmazonLightsail}.\footnote{The source code of our implementations is available at \nolinkurl{https://github.com/shokri-matin/Fed-eGlass}} We measured the latency for training a model based on our proposed framework and investigated its scalability. We repeated our experiments based on three scenarios. In each scenario, we have a different number of data-holder parties located at various locations on Earth. Figure \ref{fig:Locations_AWS} shows the geographical locations of the mediator and data-holder parties.

Table \ref{tab:AWs} explains the three scenarios for our experiments in this paper. In all scenarios, the mediator is located in Sweden. In the first scenario, we have three data-holder parties located in England, France, and Germany. In the second scenario, we have five data-holder parties located in Canada, England, France, Germany, and India. In the third scenario, we have ten data-holder parties located in Canada, the United States, England, France, Germany, India, Singapore, South Korea, Japan, and Australia. The machines in all locations run Ubuntu 20.04 and include a 1-core 2.40 GHz CPU and 1 GB RAM, hence similar resources as existing mobile devices. 

We measure the latency for one epoch of training a model based on our proposed framework for the aforementioned scenarios. The results of our experiments are reported in Table \ref{tab:AWs}. The reported results are the average results for 100 epochs of training. The latency of each epoch includes the latency of performing the required calculations on all involved machines, the mediator and data-holders, the latency of sending and receiving global parameters that the mediator shares with other parties, and the latency of communication for the secure aggregation process. 

We performed two sets of experiments. In the first set of experiments, all data holders hold equal sizes of training data. As Table \ref{tab:AWs} shows, in this setting, the average latency for one epoch of training a model takes 0.3178 s, 0.6281 s, and 1.1428 s for Scenarios 1, 2, and 3, respectively. The results show how the distance between the mediator and data holders can affect the latency for training a model.

In the second set of experiments, we divided the training data among parties in each scenario. Therefore, in these experiments, parties in Scenario 1 hold more training data compared to Scenario 2, and parties in Scenario 2 hold more data compared to Scenario 3. As Table \ref{tab:AWs} shows, in Scenario 1, one epoch of training using our framework takes 1.4366 s on average. For Scenarios 2 and 3, one epoch of training takes 0.6726 s and 1.1428 s on average, respectively. 
In Scenario 1, due to the size of training data, which is larger than the size of training data in other scenarios, the latency is the highest on average. This is despite the fact that the geographical locations of data holders in this scenario are closer to the mediator compared to other scenarios. In Scenarios 2 and 3, the distance between data holders and the mediator increases. This leads to an increase in the latency for communication, and hence in Scenario 3, although each party holds less training data, the latency is higher than the latency in Scenario 2.

\begin{table}[t]
\caption{The scenarios for our experiments on Amazon's AWS cloud and the latencies for one epoch training (average of 100 epochs)}
\label{tab:AWs}
\resizebox{\columnwidth}{!}{
\setlength{\tabcolsep}{2pt}
\begin{tabular}{lccc}
\Xhline{3\arrayrulewidth}
\multicolumn{1}{c}{}    & \textbf{Scenario 1} & \textbf{Scenario 2}     & \textbf{Scenario 3}                                                               \\ \Xhline{3\arrayrulewidth}
Number of Data Holders  & 3          & 5              & 10                                                                       \\ \hline
Mediator Location       & SE         & SE             & SE                                                                       \\ \hline
Data Holders' Locations & EN,FR,DE      & \begin{tabular}[c]{@{}c@{}}EN,FR,DE,\\CA,IN \end{tabular} & \begin{tabular}[c]{@{}c@{}}EN,FR,DE,\\CA,IN,US, \\SG,KR,JP,AU\end{tabular} \\ \hline

\begin{tabular}[c]{@{}l@{}}Latency on AWS Platform\\ (equal size of training data)\end{tabular} & 0.3178 s        & 0.6281 s            & 1.1428 s                                                            \\\hline

\begin{tabular}[c]{@{}c@{}}Latency on AWS Platform\\ (training data divided \\ equally among parties)\end{tabular} & 1.4366 s        & 0.6726 s            & 1.1428 s      \\
\Xhline{3\arrayrulewidth}
\end{tabular}}
\end{table}

\section{Conclusions}
\label{sec:Conclusion}
In this paper, we have focused on \gls{fl} in the context of health monitoring using resource-constrained mobile-health technologies, involving sensitive personal/medical data. 
We have proposed a privacy-preserving edge \gls{fl} framework for resource-constrained mobile-health technologies over the \gls{iot} infrastructure. We have evaluated our proposed framework extensively and provided the implementation of our framework on Amazon's AWS cloud platform for epilepsy monitoring, showing the relevance of our proposed framework. 

\color{black} In this work, we assume that the devices participating in the \gls{fl} process have similar resources, e.g., e-Glass systems used by the patients are one single type of device. However, in the real-world IoT applications, the mobile devices are generally heterogeneous. The proposed scheme in this article may be extended to the settings with heterogeneous resource-constrained devices assuming synchronous updates to the federated model. In such settings, the latency of each round of \gls{fl} updates among the heterogeneous resource-constrained devices will inevitably be determined by the slowest device participating in this round. In other words, while the proposed scheme can still be adopted in the context of heterogeneous devices, it has not been designed for such settings. In our future work, we will investigate privacy-preserving edge \gls{fl} designed specifically for heterogeneous devices with heterogeneous computing power/resources.\color{black}

\color{black}
Finally, in this work, we have mainly focused on the honest-but-curious privacy model. In our future work, we plan to extend this work towards other attack models, e.g., poisoning, Byzantine, and backdoor attacks \cite{xie2020fall, tolpegin2020data, bagdasaryan2020backdoor, lyu2022privacy, tramer2022truth}.
\color{black}

\bibliographystyle{elsarticle-num}

\bibliography{cas-refs}

\bio{Amin_A}
Amin Aminifar received his Ph.D. degree in computer science from the Western Norway University of Applied Sciences, Bergen, Norway, in 2022. He is currently a ZITI postdoctoral fellow at the Institute of Computer Engineering (ZITI), Heidelberg University, Germany. His current research interests include privacy-preserving federated learning and distributed machine learning with applications in the healthcare domain, as well as machine learning in resource-constrained devices in the healthcare domain.
\endbio


\bio{Matin_S}
Matin Shokri received his M.Sc. degree in computer engineering from the K. N. Toosi University of Technology, Tehran, Iran, in 2017. He is currently working as a senior data scientist at mobile phone network operator company Rightel. His research interests include deep learning, reinforcement learning, and ensemble methods.
\endbio

\vspace{1.2cm}

\bio{Amir_A}
Amir Aminifar is Assistant Professor in the Department of Electrical and Information Technology at Lund University, Sweden. He received his Ph.D. degree from the Swedish National Computer Science Graduate School (CUGS), Sweden, in 2016. During 2016-2020, he held a Scientist position in the Institute of Electrical Engineering at the Swiss Federal Institute of Technology (EPFL), Switzerland. His research interests are centered around federated and edge machine learning for Internet of Things (IoT) systems, intelligent mobile-health and wearable systems, and health informatics.
\endbio

\end{document}